\documentclass[11pt]{article}

\usepackage[preprint]{acl}
\usepackage{siunitx, graphicx}
\usepackage{times}
\usepackage{latexsym}
\usepackage{hyperref}

\usepackage{booktabs}
\usepackage{makecell} 
\usepackage{pgffor} 
\usepackage{subcaption}
\usepackage{graphicx}

\usepackage{xcolor}
\usepackage{soul}

\usepackage[T1]{fontenc}

\usepackage[utf8]{inputenc}

\usepackage{microtype}
\usepackage{enumitem}
\usepackage{inconsolata}
\usepackage{graphicx}
\usepackage{float}
\usepackage{tabularx}
\usepackage{booktabs}
\usepackage{array}
\usepackage{multirow}
\usepackage{fontawesome5}
\usepackage{hyperref} 

\definecolor{best}{HTML}{DFF0D8}    
\definecolor{worst}{HTML}{F2DEDE}   

\usepackage{booktabs}
\usepackage{makecell}
\usepackage{array}
\usepackage{siunitx}
\usepackage{placeins} 
\usepackage{tabularx, booktabs, makecell}
\usepackage{tabularx, booktabs, makecell}
\usepackage[svgnames]{xcolor}
\usepackage{enumitem}
\usepackage[most]{tcolorbox}

\usepackage{booktabs}
\usepackage{tabularx}   
\usepackage{makecell}   
\usepackage[table]{xcolor} 

\usepackage{graphicx}
\usepackage{subcaption}   
\usepackage{placeins}     
\usepackage{enumitem}     

\usepackage[normalem]{ulem} 

\definecolor{eqcolor}{HTML}{D1FAE5}   
\definecolor{discolor}{HTML}{FFE4E6}  
\definecolor{reccolor}{HTML}{FEF3C7}  
\definecolor{attcolor}{HTML}{DBEAFE}  
\definecolor{newcolor}{HTML}{F3E8FF}  

\DeclareRobustCommand{\hleq}[1]{{\sethlcolor{eqcolor}\hl{#1}}}
\DeclareRobustCommand{\hldis}[1]{{\sethlcolor{discolor}\hl{#1}}}
\DeclareRobustCommand{\hlrec}[1]{{\sethlcolor{reccolor}\hl{#1}}}
\DeclareRobustCommand{\hlatt}[1]{{\sethlcolor{attcolor}\hl{#1}}}
\DeclareRobustCommand{\hlnew}[1]{{\sethlcolor{newcolor}\hl{#1}}}

\title{Retell, Reward, Repeat: Reinforcement Learning for Narrative Theory-Informed Counterfactual Storytelling}

\author{
  \textbf{David Y. Liu}$^{1}$\thanks{Corresponding Author: \texttt{david.liu2@unsw.edu.au}} \quad
  \textbf{Xanthe Muston}$^{2}$ \quad
  \textbf{Dipankar Srirag}$^{1}$ \\
  \textbf{Aditya Joshi}$^{1}$ \quad
  \textbf{Paul Dawson}$^{2}$ \quad
  \textbf{Sebastian Sequoiah-Grayson}$^{1}$ \\[1ex]
  $^{1}$School of Computer Science and Engineering, UNSW Sydney \\
  $^{2}$School of the Arts and Media, UNSW Sydney \\
}

\begin{document}
\maketitle
\begin{abstract}
Counterfactual storytelling exposes LLM shortcomings in causal reasoning, where few-shot prompting and SFT fail to produce logical and rational narratives. This case study introduces Retell, Reward, Repeat (RRR), a narrative theory-informed RL pipeline that teaches causal story structure. We extend an existing counterfactual stories dataset with human-annotated stages of narrative equilibrium, synthesising Structuralist Narratology with scalar narrativity. An aligned LLM-as-judge automates this framework to output reward signals, teaching the policy LLM to generate texts with causal narrative structure. Evaluations show that RRR-trained 7/8B LLMs outperform few-shot and SFT baselines in logic, rationality, and completeness, also aligning with a blind human preference subset. RRR offers a linguistically grounded, cost-effective post-training mechanism for LLM storytelling using a small, query-only dataset, showing the value of integrating linguistic theories into contemporary NLP.

\end{abstract}

\section{Introduction}

     Storytelling differs from many Natural Language Processing (NLP) tasks due to the lack of a universally preferable output. While statistical metrics based on reference outputs have been a norm in NLP (e.g., BLEU \cite{papineni2002bleu}, ROUGE \cite{lin2004rouge} and BERTScore \cite{ZhangKWWA20}), they align poorly/negatively with human judgements of Automatic Story Generation (ASG) outputs \citep{netisopakul2023comparison}. 
     
     Existing large language model (LLM) post-training paradigms such as Supervised Fine-tuning (SFT) and Reinforcement Learning from Human Feedback (RLHF) restrict the diversity of story generations \citep{lu2025ai,sui2026llms}, where models converge towards clichéd and contextually unnecessary linguistic styles\footnote{ASG literature sometimes defaults to reductive universal criteria such as suspense and entertainment. However, soothing bedtime stories are not suspenseful, and stories depicting genocide are not entertaining.} \citep{chakrabarty2025can}. 
    
    \begin{figure}[h]
        \centering
        \includegraphics[width=0.8\linewidth]{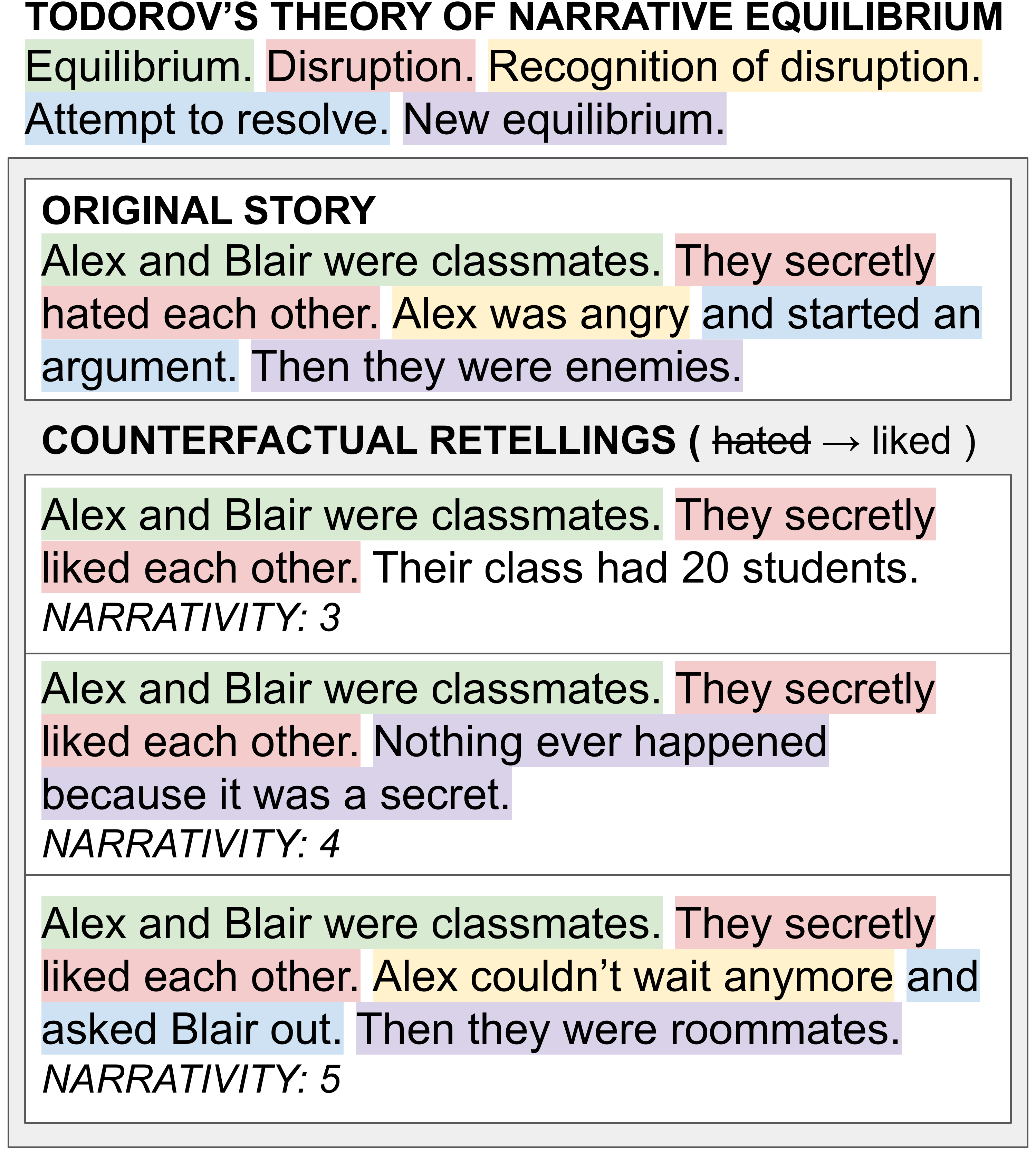}
        \caption{\citet{todorov19712}'s narrative theory applied to \citet{qin-etal-2019-counterfactual}'s counterfactual storytelling task. Narrativity measures the presence of narrative structure.}
        \label{fig:application}
    \end{figure}

    We move towards a more narrative-aligned LLM post-training by redefining how we measure storytelling. When we formulate the problem of ASG as selecting and representing a sequence of events that \textbf{(A)} can be told as a story and \textbf{(B)} make a good story \citep{li2013story}, we can see that current frameworks \citep{chhun2022human, chakrabarty2024art} disproportionately prioritise \textbf{(B)}. This drives an impossible search for ‘universally good storytelling’, as evidenced by conflicting subjective human evaluations of ASG outputs \citep{marco2025reader} and the absence of standardised benchmarks or post-training approaches \citep{liu2026narrative}. 

    \begin{figure*}[ht]
        \centering
        \includegraphics[width=0.9\linewidth]{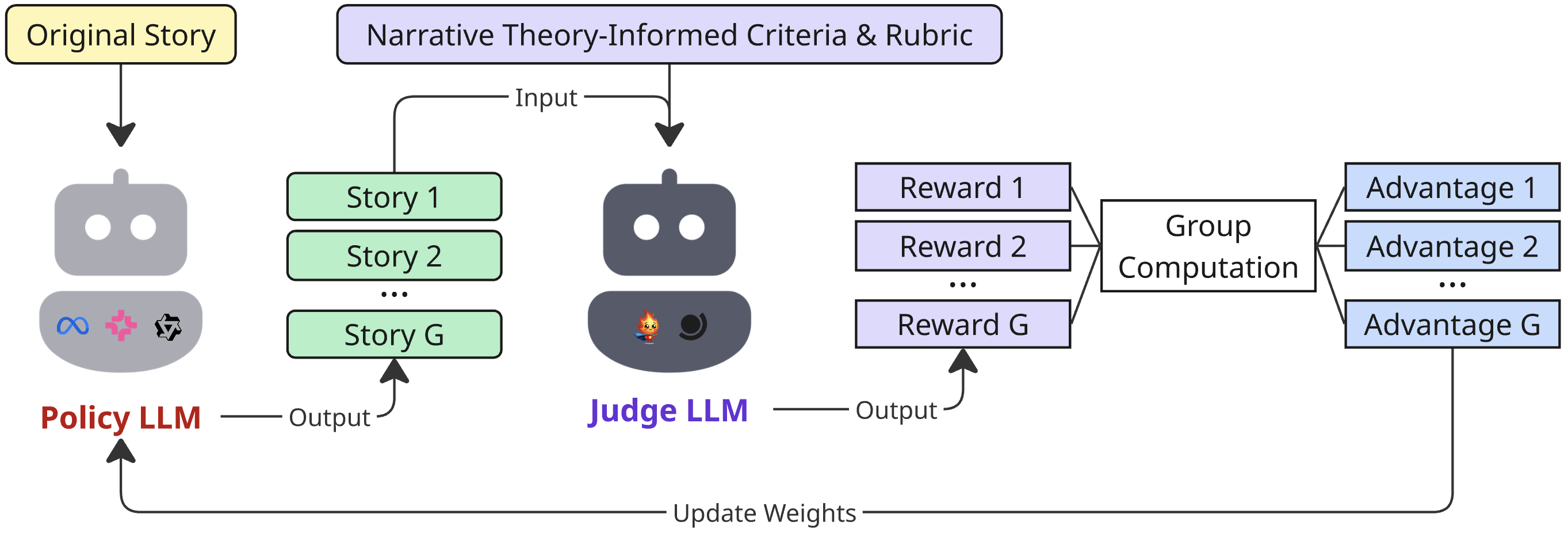}
        \caption{The Retell Reward Repeat (RRR) post-training pipeline uses d-RLAIF \citep{lee2024rlaif} with a narrative theory-informed LLM-as-judge to generate the reward signal for GRPO \citep{shao2024deepseekmath}. 
        \label{fig:drlaif}}
    \end{figure*}
    
    Proposing an alternative training signal, we shift our priority from \textbf{(B)} to \textbf{(A)} and seek to measure the text's narrativity, fundamental qualities shared by all narratives, which provides a stylistically agnostic metric across diverse storytelling forms:
    

    \begin{description}[style=unboxed,leftmargin=0cm,labelindent=0cm,itemsep=0cm,parsep=0cm] 
        \item[RQ1:] How can narrativity in LLM outputs be measured with a set of definable properties?
        \item[RQ2:] How can post-training align LLM outputs with narrativity and other storytelling criteria? 
    \end{description}
    
    

    Open-ended prompting can be permissive and unreliable when evaluating storytelling abilities, where LLMs can easily reproduce memorised texts seen during training; introducing constraints, such as counterfactuals, forces LLMs to select a sequence of logically and rationally connected causal events, which empirically increases task difficulty \citep{atmakuru2024cs4}. Accordingly, we select the counterfactual story reasoning and generation task introduced by \citet{qin-etal-2019-counterfactual} (Figure \ref{fig:application}) for this ASG case study. Our contributions are as follows:
    
    \begin{itemize}[noitemsep, topsep=0pt, leftmargin=*]
        \item We construct an annotation scheme that synthesises \citet{todorov19712}'s Structuralist Narratology with \citet{prince2003dictionary}'s scalar narrativity, and additionally define three criteria (Logical, Rational, Complete\textsubscript{N}) for counterfactual story retelling. 
    
        \item We curate a dataset of 200 counterfactual retellings with textual features tagged and evaluated by three human annotators. Human-authored stories show higher narrativity and satisfy the three criteria more consistently than LLM-generations. Additionally, we analyse the inter-annotator agreement/correlation to empirically show the usefulness and limitations of Todorov's theory as ASG criteria. 
        
        \item We use the human annotations to benchmark three LLM-as-judge candidates and identify Gemini-3-Flash~\citep{google2025gemini3flash} as the evaluator most aligned with human judgements.
    
        \item We introduce a theory-informed `Retell Reward Repeat' pipeline (RRR; Figure \ref{fig:drlaif}) and show that 7B/8B LLMs trained with RRR achieve the highest narrativity, and satisfy the three retelling criteria more consistently than LLMs trained with SFT on human-authored ground truths.

        \item Implementation and dataset are available on GitHub: \href{https://github.com/daavidliu/Retell-Reward-Repeat}{\faGithub\ \url{https://github.com/daavidliu/Retell-Reward-Repeat}}.

    \end{itemize}

    By engaging with debates in literary theory, our study synthesises linguistically grounded narrative frameworks to confront the fundamental challenges of aligning LLM post-training with storytelling abilities. While existing evaluation criteria for storytelling remain valuable, our work demonstrates the necessity of exploring alternative pathways to overcome limitations in current training paradigms. 


\section{Structuralist Narratology \& Narrativity}
\label{sec:prelim}

    \citet{todorov1969grammaire} coined the term \textit{narratologie} to describe a systematic study of narrative which seeks to discover the abstract structure and essential textual features that universally and exclusively define a narrative. Structuralist Narratology treats an individual story as just one possible manifestation of narrative's underlying universal structure. 
    
    \citet{todorov19712}'s plot-focused Theory of Narrative Equilibrium claims that narrative's underlying universal structure can be represented by sequences of causally and contextually connected transformations, categorisable into five types of narrative stages. We illustrate these `Todorovian stages' in Figure~\ref{fig:application} where each stage is colour‑coded by type.

    While Classical Narratology treated a text's narrativeness as a boolean property, Narratology has increasingly embraced the concept of scalar narrativity \citep{prince2003dictionary}: narrativeness as a matter of degree rather than an absolute. \citet{todorov19712} implied this continuum in his original work: ``one cannot imagine a narrative which does not contain at least one part of [the five equilibrium stages].''
    
    We synthesise these theories to operationalise a measurement of narrativity to evaluate the causal narrative structure of a counterfactual retelling (Figure \ref{fig:application}), where a text containing five types of Todorovian stages is considered to exhibit higher narrativity than a text containing four or three.

    By testing Todorov's assumption that narrativity relies on universal textual features, our case study engages with broader debates in literary theory. Post-structuralists like \citet{barthes1967death} argue that interpretation relies on more than just textual features, as narratives exist within fluid cognitive/cultural structures inhabited and contested by ideas. 
    
    \textbf{RQ1} grounds the issue empirically by investigating how narrativity itself can be measured from textual features. Ultimately, our annotation process serves to test the boundaries of Todorov's structuralist framework.

\section{Story Annotation}
    \citet{qin-etal-2019-counterfactual} show that for the task of rewriting the ending of a short story after a provided `Counterfactual'\footnote{The `Counterfactual' is a counterfactual antecedent while the `Edited Ending' is a counterfactual consequent.} event replaces the `Initial' event (Table~\ref{fig:application}), an AI-generated output's similarity\footnote{Measured using BLEU \cite{papineni2002bleu}, ROUGE \cite{lin2004rouge} \& BERTScore \cite{ZhangKWWA20}.} to the human-authored ground truths correlates weakly and sometimes negatively with human given ratings of coherence and relevance. To overcome the unreliability of ground truth based evaluation, our case study applies Todorov's theory to provide a linguistically grounded approach to automatic metrics for training and evaluation.

    The functional purpose of the annotation dataset is solely evaluative: i.e., to test agreement between human annotators and LLM-evaluators to inform our decisions in selecting LLM-as-judge models for training and evaluation. The annotated data is not directly used for model training. Additionally, through quantitatively and qualitatively examining the inter-annotator agreement, we can empirically observe whether and how we can measure narrativeness using definable conditions (\textbf{RQ1}).

\subsection{Curation}
\label{subsec:curation}

    We curate a pool of human and AI-generated retellings to reflect the distribution of texts the LLM-evaluator will see during the reinforcement learning process and final evaluation. 
    
    We start the curation process from the first 3000 items in the TimeTravel supervised training split \citep{qin-etal-2019-counterfactual}, where each item (Table \ref{tab:example_outputs}) provides a `query' consisting of the premise, initial state, original ending, and counterfactual, and a `response' which is the edited ending (human-authored ground truth). For each query, we produce three AI-generated responses using Llama-3.1-8B-Instruct \citep{grattafiori2024llama}, Olmo-3-7B-Instruct \citep{olmo2025olmo}, and Qwen-3-8B \citep{yang2025qwen3} with thinking disabled.

    Because the three instruction-tuned LLMs typically produce similar responses to the same query, we introduce a filter aiming to diversify the data. We do not include the human authored ground truth in the calculation of our diversity metric to avoid biasing our selection towards `harder' queries where the LLM output is distant from the ground truth, even if it is only a weak correlation. 
    
    For each query's three AI-generated responses, we use DeBERTa-v3-small \citep{He2023DeBERTaV3ICLR} to compute all pairwise distances (defined as $1 - \mathrm{BERTScore}_{\mathrm{F1}}$) and define the diversity metric as $\operatorname{diversity} = \min(\text{distances}) \times \operatorname{mean}(\text{distances})$. We then select the 50 queries with the highest diversity and, for each, retain four responses (the three LLM generations plus the human-authored ground truth), yielding an annotation set of $n = 200$.

\begin{table*}[h]
    \renewcommand{\arraystretch}{0.9}
    \footnotesize
    \centering
    \setlength{\tabcolsep}{3pt}
    \begin{tabularx}{\textwidth}{@{}X@{}} 
    \toprule
    \textbf{Narrativity} (\hleq{Equilibrium}→\hldis{Disruption}→\hlrec{Recognition}→\hlatt{Attempt}→\hlnew{New Equilibrium})\\ 
    \textbf{A:}
    (4)
    \hleq{Evan was ready to buy himself a bicycle.} 
    \hldis{He went to the store but forgot his money.} 
    \hlatt{Then he looked around some more, trying to find something else he could afford or a way to get home.}
    \\
    \textbf{B:} 
    (4)
    \hldis{Evan was ready to buy himself a bicycle.} 
    \hlatt{He went to the store} 
    \hldis{but forgot his money.} 
    \hlnew{Then he looked around some more, trying to find something else he could afford or a way to get home.}
    \\
    \textbf{C:} 
    (3)
    \hldis{Evan was ready to buy himself a bicycle.} 
    \hlatt{He went to the store but forgot his money.} 
    Then he looked around some more, trying to find something else he could afford or a way to get home.
    \\
    \toprule 
    
    \textbf{Is the story logical?} \\ 
    \textbf{A:} (3) I interpret ‘but forgot his money’ as Evan forgetting a large sum of money intended for buying a bicycle. Evan may still have the usual money he carries every day, and is able to logically find something else he could afford. \\
    \textbf{B:} (3) The edited text is logical. Forgetting his money and then looking for an alternative or a way home is internally consistent. \\
    \textbf{C:} (1) There is a logical impossibility in the story because: 1. Evan forgets his money, but 2. proceeds to look around the store for something he can “afford,” which would be nothing. \\
    \bottomrule
    \end{tabularx}
    \caption{An example of 3 human annotators (A/B/C) applying our framework. More examples are in Appendix \ref{sec:inter_annotator_diff}.}
    \label{tab:inter_annotator_bicycle}
\end{table*}

\subsection{Scheme}
\label{subsec:scheme}
    
    We operationalise our scalar measurement of narrativity as the number, $s$, of recognisable Todorovian stages in a text (Table~\ref{fig:application}). Since disruption is essential for narrative transformation, any text lacking a disruption is assigned a score of `1'. For all other cases, the narrativity score is defined as $\mathrm{Narrativity} = \min(s + 1,\, 5)$. Our conditions only require four recognisable narrative stages to achieve the maximum score since we do not wish to penalise stories which begin \textit{in medias res} or end on a cliffhanger (Appendix \ref{subsec:variations}). 
     
     In addition to scalar narrativity, we define specific criteria to measure quality for the counterfactual story retelling task (\textbf{RQ2}), designed to align with the evaluation framework originally proposed by \citet{qin-etal-2019-counterfactual}. While their metrics evaluate a model's ability to respect counterfactual changes, retain original premise details, and preserve plot relevance, we adapt these dimensions into our Logical, Rational, and Complete\textsubscript{N} criteria, respectively. Our approach differs by grounding these definitions in Todorov's narrative theory wherever possible:

     
    \begin{description}[style=unboxed,leftmargin=0cm,labelindent=0cm,itemsep=0cm,parsep=0cm] 
        \item[Logical:] The text is free from inconceivable scenario(s) that contradict the text's internal logic.
        \item[Rational:] The text can be rationally interpreted using Todorovian stage(s) in its entirety without contextual/causal disconnection.
        \item[Complete\textsubscript{N}:] The text includes all key Todorovian stage(s) from the original that ensure the text is as narratively complete as the original.
        \item[min\textsubscript{LRC}:] Since any desirable story needs to satisfy all 3 of the aforementioned criteria, we use \( \text{min}(\text{Logical}, \text{Rational}, \text{Complete}_{N}) \) to represent a text's overall quality instead of the average. This reflects the principle that failing even one criterion makes the story completely undesirable.
    \end{description}
    
    
\subsection{Human Annotation}
    \label{subsec:inter_annotator}

    \begin{table}[h]
        \centering
        \footnotesize
        \setlength{\tabcolsep}{3pt} 
        \begin{tabularx}{\columnwidth}{Xcccc}
        \toprule
        Annotator & Logical & Rational & Complete\textsubscript{N} & Narrativity \\
        \midrule
        A & 0.31 & 0.53 & 0.46 & 0.43 \\
        B & 0.24 & 0.37 & 0.38 & 0.53 \\
        C & 0.17 & 0.46 & 0.37 & 0.49 \\
        \midrule
        \textbf{Average} & \textbf{0.24} & \textbf{0.46} & \textbf{0.40} & \textbf{0.48} \\
        \bottomrule
        \end{tabularx}
        \caption{Correlation (Spearman's $\rho$) of each annotator vs the median of the other two ($n=200$). All $p < 0.05$.}
        \label{tab:human-annotator-spearman}
    \end{table}
    
    Three human annotators, blind to whether the stories are sourced from humans or LLMs, annotate the texts using a tag-and-evaluate process: (1) Tag parts of text with Todorovian stages. The variety of tags is automatically counted, and \textit{post hoc} computed into the 1-5 Narrativity score described in Section \ref{subsec:scheme}, then (2) use 3-point Likert scales to rate the  Logical/Rational/Complete\textsubscript{N} criteria. Additional instructions and examples given to the annotators are included in Appendix \ref{sec:annotator_instructions}.

    Story quality and narrativity arise from subjective interactions between readers and texts \citep{marco2025reader}. This inherent subjectivity, consistent with reader-centric post-structuralist literary theory \citep{barthes1967death}, accounts for the fair to moderate agreement observed from annotators (Table \ref{tab:human-annotator-spearman}). Textual features capture only one dimension of narrative understanding, which is ultimately completed by the reader's cognitive processing.

    After acquiring some additional rationales from the annotators, a qualitative review of the disagreements (Table \ref{tab:inter_annotator_bicycle}) shows that annotators often formed differing yet valid interpretations of the same text. Readers frequently disagreed on structural starting points—such as whether a narrative begins in a state of equilibrium or in disruption—and applied subjective assumptions to resolve ambiguities.

    For context, \citet{chhun2022human} previously used ICC2k to report agreement between 3 human annotators with an average of 0.40 across 6 criteria: relevance (0.48), coherence (0.29), empathy (0.34), surprise (0.28), engagement (0.46) and complexity (0.56). In comparison, the ICC2k of our criteria--Narrativity (0.64), Logical (0.36), Rational (0.67), Complete\textsubscript{N} (0.58) and min\textsubscript{LRC} (0.56)—align with the upper range of expected inter-annotator agreement when subjectively evaluating narratives only based on their textual features.  


    When we use the collected annotations to rank the authors (Human/Olmo/Qwen/Llama) by their average received ratings, the blind evaluations consistently yield the same rankings of authors across all four criteria: Logical, Rational, $\text{Complete}_N$, and Narrativity (see Appendix Table \ref{tab:ranking_200}). Human authored ground truths are consistently rated higher in both narrativity and overall quality than LLM generations. These human annotations show that our criteria establish a meaningful, non-trivial benchmark for the counterfactual story retelling task by finding a distinct capability gap between LLMs and human-authors (Table \ref{tab:example_outputs}).

    


    \begin{table*}[h]
        \footnotesize
        \centering
        
        \noindent
        \begin{tabularx}{\textwidth}{@{} l X @{}}
        \toprule
        \rowcolor{black!5}\multicolumn{2}{@{} l}{\textbf{(query)} Original Story + \textbf{Counterfactual}} \\
        \midrule
        Premise         & Kimmy was on a trip to Paris. \\
        Initial         & Her plane had just landed and she and her classmates her \textit{[sic]} excited. \\
        Original Ending & They couldn’t wait to start seeing museums and churches! They spent lots of time shopping and seeing historical sites. Kimmy was very excited. \\
        \textbf{Counterfactual}  & Kimmy’s plane crashed over the ocean and killed everyone on board. \\
        \bottomrule
        \end{tabularx}
        
        \vspace{6pt}
        
        \noindent
        \begin{tabularx}{\textwidth}{@{} l X @{\hspace{1.2em}} c @{\hspace{0.8em}} c @{\hspace{0.8em}} c @{\hspace{0.8em}} c @{\hspace{1.2em}} p{4.2cm} @{}}
        \rowcolor{black!5}\multicolumn{7}{@{} l}{\textbf{(response)} Premise → \sout{Initial} → Counterfactual → \textbf{Edited Ending}} \\
        \midrule
        \textbf{Source/Model} & \textbf{Counterfactual Story} & \textbf{L} & \textbf{R} & \textbf{C} & \textbf{min\textsubscript{LRC}} & \textbf{Our Comment} \\
        \midrule
        Ground Truth & Kimmy was on a trip to Paris. Kimmy's plane crashed over the ocean and killed everyone on board. Kimmy had just wanted to see the museums and churches! Her parents spent a lot of time planning her funeral. Kimmy was dead. & 3 & 3 & 3 & 3 & The scenarios are logical, the sequence is rationally connected, and the story feels complete. \\
        \midrule
        \addlinespace
        Zero-Shot    & Kimmy was on a trip to Paris. Kimmy's plane crashed over the ocean and killed everyone on board. They never made it to Paris. Kimmy's plane crashed over the ocean and there were no survivors. & 3 & 1 & 3 & 1 & The crash is described twice. Irrationally repetitive. \\
        \addlinespace
        Few-Shot     & Kimmy was on a trip to Paris. Kimmy's plane crashed over the ocean and killed everyone on board. Kimmy's plane crashed over the ocean and killed her. & 1 & 3 & 1 & 1 & The scenario of Kimmy's plane crashing and killing her again is illogical. The story does not feel complete either. \\
        \addlinespace
        SFT & Kimmy was on a trip to Paris. Kimmy's plane crashed over the ocean and killed everyone on board. They couldn’t wait to start seeing museums and churches! They spent lots of time shopping and seeing historical sites. Kimmy was very excited & 1 & 1 & 3 & 1 & The contradiction of the plane crash and then the Paris visit is illogical. There is no rational causal/contextual connection between the events. \\
        \addlinespace
        RRR\textsubscript{N} & Kimmy was on a trip to Paris. Kimmy's plane crashed over the ocean and killed everyone on board. The plane wreckage was found at the bottom of the ocean, a tragic reminder of the devastating crash that took Kimmy and her classmates' lives. & \textbf{3} & \textbf{3} & \textbf{3} & \textbf{3} & The scenarios are logical, the sequence is rationally connected, and the story feels complete. \\
        \bottomrule
        \end{tabularx}
        \caption{Example test outputs from models based on Llama-3.1-8B-Instruct. Ratings are given by LLM-as-judge (Gemini-3-Flash): L=Logical, R=Rational, C=Complete\textsubscript{N}, M=min\textsubscript{LRC}. Additional examples in Appendix \ref{sec:example_outputs}.}
        \label{tab:example_outputs}
    \end{table*}
    
\subsection{LLM-as-Judge}
\label{sec:llm_as_judge}
    \begin{table}[H]
        \centering
        \footnotesize
        \setlength{\tabcolsep}{3pt} 
        \begin{tabularx}{\columnwidth}{Xcccc}
        \toprule
        Model & Logical & Rational & Complete\textsubscript{N} & Narrativity \\
        \midrule
        Selene & $-0.05^{\dagger}$ & 0.31 & 0.23 & 0.39 \\
        Prometheus  & 0.24 & 0.29 & 0.33 & 0.32 \\
        Gemini & \textbf{0.34} & \textbf{0.39} & \textbf{0.52} & \textbf{0.50} \\
        \bottomrule
        \end{tabularx}
        \caption{Correlation (Spearman's $\rho$) of LLM-as-Judge vs the median of 3 human annotators ($n=200$). All $p < 0.05$ except Selene Logical ($^{\dagger}p = 0.47$).}
        \label{tab:llm-judge-spearman}
    \end{table}
    
    We test 3 LLM-as-judges: Selene-1-mini-8B \citep{alexandru2025atla}, M-Prometheus-14B \citep{pombal2025m} and Gemini-3-Flash \citep{google2025gemini3flash}. Selene and Prometheus are open-weight models that specialise in evaluation tasks, while Gemini is chosen as a state-of-the-art (SOTA) proprietary model. We compute Spearman's $\rho$ correlations between each LLM-judge and the median of three human annotators (Table \ref{tab:llm-judge-spearman}). 
    
    Gemini achieves an average correlation of 0.44, which is comparable to the average human annotator—the correlation between an average human annotator and the median of the remaining two is 0.40 (Table \ref{tab:human-annotator-spearman}). In contrast, open-weight models achieve lower correlations, 0.25 for Selene and 0.30 for Prometheus. These results show that Gemini serves as a good proxy for human judgement. We select Gemini for our final evaluation as it is the most reliable and human-aligned.

\subsubsection{Comparison with Previous Metrics}
    
    \citet{qin-etal-2019-counterfactual}'s dataset and task are evaluated in the existing literature \citep{liu-etal-2023-magic,liu2025eliciting} using automatic-metrics--BLEU-4, ROUGE-L and BERTScore--out of which only BERTScore has a positive correlation with all the original 3 human-annotated criteria. In comparison, our chosen Gemini generated automatic metric minLRC and the 3 human rated criteria have an average Pearson’s correlation of 0.30 (vs BERTScore’s 0.17). Therefore, our automatic evaluation is an improvement.

\setlength{\tabcolsep}{3pt} 
\renewcommand{\arraystretch}{0.9}

\section{Retell, Reward, Repeat (d-RLAIF)}
    As shown in Figure \ref{fig:drlaif}, we adapt \citet{wei-etal-2025-igniting}'s implementation of d-RLAIF to use an LLM-as-judge to transform an input story into a theory-informed reward signal. We use GRPO \cite{shao2024deepseekmath} and LoRA \cite{hu2022lora} to optimise the RRR training process using TimeTravel's unsupervised training split. To study the effects of our RRR training on a wide range of models, three instruction-tuned models (Llama-3.1-8B, Qwen-3-8B, and Olmo-3-7B) are chosen as policy models for generating retellings.

    The policy model generates 16 responses for each query, which are then given to the LLM-as-judge to evaluate and generate 16 reward scores. The relative advantages are then calculated from the reward scores and used to update the weights (Figure \ref{fig:drlaif}) of the policy model's LoRA adapter.

\begin{figure}[H]
        \centering
        \includegraphics[width=\columnwidth]{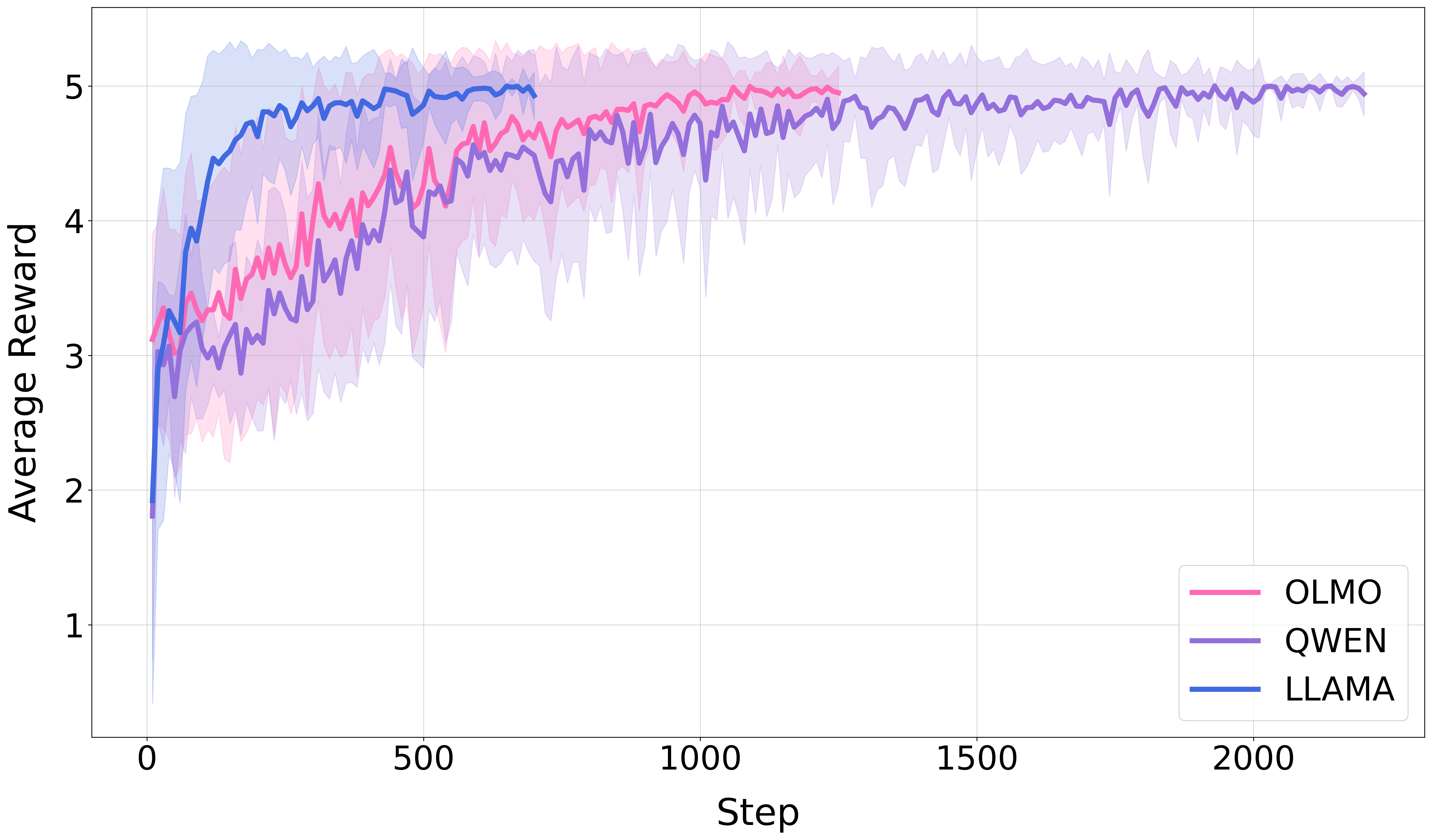}
        \caption{Mean and standard deviation of the reward signal R\textsubscript{N} generated by Selene-1-mini during d-RLAIF.}
        \label{fig:rn_training_graph}
    \end{figure}

    We reserve Gemini-3-Flash--our most human-aligned LLM-judge (Section \ref{sec:llm_as_judge})--for final evaluation rather than training, to keep the training reward signal and the evaluation metric independent and to prevent reward hacking. Between the two remaining evaluators, we assign each to the reward signal matching its relative strength in Table \ref{tab:llm_as_judge_vs_human}: Selene achieves the second-highest correlation with human judgment on narrativity specifically, so we use it to generate the narrativity-based reward signal $R_N$ (Figure \ref{fig:rn_training_graph}). For comparison, Prometheus achieves the second-highest correlation on overall story quality, so we use it to compute the alternative reward signal $R_O$ (`overall') using a prompt with criteria identical to the $\min_{\text{LRC}}$ score. Although this configuration aligns less closely with human judgements than the three-inference aggregated $\min_{\text{LRC}}$ score (Appendix Table \ref{tab:llm_as_judge_vs_human}), we consider the minor performance trade-off acceptable given the $\frac{2}{3}$ reduction in token costs. We detail additional ablation experiments on the reward design in Appendix \ref{appendix_training}.
    
    


    We apply early stopping once training reaches a reward plateau, defined as 200 consecutive optimiser steps for which the average reward remains within 0.1 of the maximum achievable reward, indicating no further measurable improvement. Additionally, we introduce a length-based penalty for retellings that exceed three times the length of the original to prevent lengthy outputs. 

\section{Evaluation}
\label{sec:results}

        \begin{table*}[t]
      \centering
      \footnotesize 
      \begin{tabular}{ll lllllll}
        \toprule
        Model & 
        Method & 
        Logical &
        Rational &
        Complete\textsubscript{N} &
        \makecell{\textbf{min}\textsubscript{LRC}} &
        Narrativity &
        BLEU-4 &
        ROUGE-L \\
        \midrule
        
        \textbf{\textsc{Ground Truths}} & Crowd Workers & 2.884 & 2.927 & 2.933 & 2.811 & 4.718 & - & - \\
        \midrule
        
        \multirow{5}{*}{\textsc{Llama-3.1-8B-Inst.}}
        & BASE                 & 2.735 & 2.385 & 2.528 & 2.080 & 4.347 & 0.340 & 0.502 \\
        & few-shot             & 2.768 & 2.615 & 2.505 & 2.213 & 4.303 & 0.306 & 0.473 \\
        & SFT                  & 2.320 & 2.501 & \textbf{2.934} & 2.260 & 4.692 & \textbf{0.810} & \textbf{0.835} \\
        & RRR\textsubscript{O}-Prometheus & \textbf{2.801} & 2.825 & 2.594 & 2.400 & 4.534 & 0.179 & 0.386 \\
        & RRR\textsubscript{N}-Selene     & 2.785 & \textbf{2.860} & 2.740 & \textbf{2.554} & \textbf{4.733} & 0.087 & 0.299 \\
        \midrule
        
        \multirow{5}{*}{\textsc{Qwen-3-8B}}
        & BASE (non-thinking)  & 2.429 & 2.132 & 2.759 & 1.940 & 4.551 & 0.572 & 0.667 \\
        & few-shot             & 2.543 & 2.219 & 2.686 & 2.006 & 4.559 & 0.492 & 0.607 \\
        & SFT                  & 2.402 & 2.532 & \textbf{2.925} & 2.317 & 4.693 & \textbf{0.790} & \textbf{0.823} \\
        & RRR\textsubscript{O}-Prometheus & 2.717 & 2.336 & 2.566 & 2.024 & 4.356 & 0.337 & 0.491 \\
        & RRR\textsubscript{N}-Selene     & \textbf{2.781} & \textbf{2.890} & 2.705 & \textbf{2.516} & \textbf{4.894} & 0.003 & 0.192 \\
        \midrule
        
        \multirow{5}{*}{\textsc{Olmo-3-7B-Inst.}}
        & BASE                 & 2.762 & 2.812 & 2.400 & 2.222 & 4.328 & 0.097 & 0.315 \\
        & few-shot             & 2.745 & 2.837 & 2.377 & 2.208 & 4.415 & 0.072 & 0.287 \\
        & SFT                  & 2.295 & 2.441 & \textbf{2.910} & 2.211 & 4.647 & \textbf{0.787} & \textbf{0.821} \\
        & RRR\textsubscript{O}-Prometheus & 2.770 & 2.857 & 2.491 & 2.325 & 4.460 & 0.090 & 0.309 \\
        & RRR\textsubscript{N}-Selene     & \textbf{2.793} & \textbf{2.893} & 2.595 & \textbf{2.434} & \textbf{4.899} & 0.002 & 0.178 \\
        \midrule
        
        \textsc{Codex} & Program Prompt & 2.452 & 2.484 & 2.758 & 2.181 & 4.514 & 0.617 & 0.695 \\
        \bottomrule
      \end{tabular}
      
      \caption{Evaluation of post-trained LLMs using the test split (n=1871) from the TimeTravel dataset using Gemini-3-Flash. For each base LLM, the highest score in each column is bolded. Significance testing in Appendix \ref{sec:significance_testing}.}
      \label{tab:story-eval-wide}
    \end{table*}
    
    We evaluate our models on TimeTravel's test split (n=1871) using the most aligned LLM-as-judge we found in Section \ref{sec:llm_as_judge}: Gemini-3-Flash. Each item in the test split contains three slightly different human-authored endings--we evaluate each and use their average for the results in Table \ref{tab:story-eval-wide}. 
    
    Human authored ground truths receive higher average ratings than LLMs when considering {min}\textsubscript{LRC}, which better represents overall quality than the average, as all desirable stories must score perfectly across the three criteria. We also use BLEU-4 and ROUGE-L to calculate the average similarity of each model's generation to the nearest-of-three human-authored endings to measure similarity to the ground truth. 
    
    For in-context learning comparisons against our methods, we also evaluate the outputs generated by \citet{liu-etal-2023-magic}'s Program-Prompt method. Additionally, we evaluate outputs generated by zero-shot and few-shot prompting (Appendix \ref{appendix-few-shot}). For all three LLMs tested, few-shot does not result in consistent or noticeable performance over zero-shot. 
    
    For post-training comparisons, we conduct SFT with LoRA using TimeTravel's supervised training split, also with early stopping at convergence--defined by when loss decreases < 0.01 for 3 consecutive training steps (Appendix Figure \ref{fig:sft_loss}). SFT achieves the highest Complete\textsubscript{N}, BLEU-4 and ROUGE-L for all models. SFT results in an increase of min\textsubscript{LRC} performance for Qwen and Llama but degrades the performance of the best-performing Olmo. Stories generated by the SFT models are highest in similarity to the human-authored ground-truths, as well as the most structurally complete when compared to the corresponding original story (measured by Complete\textsubscript{N}) from the query (Table \ref{tab:example_outputs}).
    

    When using RRR, we find that d-RLAIF with Prometheus's R\textsubscript{O} improves the overall performance of all tested models, whereas Selene's narrativity-based reward signal R\textsubscript{N} reliably results in the best min\textsubscript{LRC} performance, second only to humans (significance testing in Appendix \ref{sec:significance_testing}). Their outputs also contain the highest narrativity and dissimilarity to human authored ground truths. 
    
\section{Discussion}
    \subsection{min\textsubscript{LRC} and Human Preference}

        \begin{table}[H]
            \centering
            \begin{tabular}{lcc}
            \toprule
               & Llama (SFT)   & Llama (RRR\textsubscript{N}) \\ \midrule
            
            Narrativity          & \textbf{4.820} & 4.580 \\
            $min\textsubscript{LRC}$          & 2.220 & \textbf{2.480} \\
            Win-Rate & 28\% & \textbf{72\%} \\ \bottomrule
            \end{tabular}
            \caption{Human preference test on a subset of 100.}
            \label{tab:human_preference}
        \end{table}

        As a sanity check to see whether our primary quality metric min\textsubscript{LRC} correlates with human preference, we instruct an additional annotator D (who is not involved in the initial annotations) to choose their preferred story on a subset of the first 50 queries and 100 shuffled responses, while being blind to which model generated it (Table \ref{tab:human_preference}).
        
        Out of 100 stories generated by Llama-SFT and Llama-RRR\textsubscript{N}, a blinded annotator D strongly prefers Llama-RRR\textsubscript{N} at 72\% of the time (Table \ref{tab:human_preference}). When looking at outputs qualitatively (Table \ref{tab:example_outputs}), it is also self-evident that stories which are logical, rational, and narratively complete are more desirable than stories which are not. This shows that, for these micro-stories, the Gemini-generated min\textsubscript{LRC} can be understood as a measurement of story quality which positively correlates with human preference.
    
    \subsection{Reinforcement Learning vs SFT}
    \label{subsec:performance}
        Intuitively, it does not make sense to evaluate a student's storytelling ability by comparing their stories with the teacher's. When we consider that humans naturally learn storytelling as a communicative practice through reinforcement learning \citep{hineline2018narrative}, our RRR pipeline using d-RLAIF is an intuitive choice for LLM post-training in the domain of Automatic Story Generation (ASG).
    
        The increases in min\textsubscript{LRC} and Narrativity across all models trained using our RRR\textsubscript{N} method (Table \ref{tab:story-eval-wide}) indicate that Todorov's abstract and theoretical narrative structure can be learned by LLMs through reinforcement learning (since Narrativity is measured by the presence of Todorovian structures) and is beneficial in generating logical, rational, and structurally complete stories--even though they differ the most from ground truths (Table \ref{tab:story-eval-wide}). 
    
        While LLMs trained through SFT achieve the highest Complete\textsubscript{N}, the highest similarity to ground truths, and the second highest Narrativity, overall these scores do not translate to higher min\textsubscript{LRC} scores or human preference. Example outputs (Table \ref{tab:example_outputs}) show that SFT models often generate high-narrativity story endings by illogically and irrationally duplicating the original ending. It appears that SFT forces LLMs to over-fit to the training data without learning the causal reasoning abilities needed for the counterfactual story retelling task.
    
        The results also highlight that, while optimising for narrativity during training (RRR\textsubscript{N}) succeeds in our case study, evaluating based on narrativity alone is insufficient. Recalling our formulation of ASG—selecting and representing a sequence of events that \textbf{(A)} form a story and \textbf{(B)} make a good story—both components are essential. While our findings demonstrate the value of prioritising \textbf{(A)} during training, evaluation must still assess the criteria \textbf{(B)} required by the specific task.

    \subsection{d-RLAIF for Storytelling}

        The lack of consistent or significant performance increases of few-shot over zero-shot (Table \ref{tab:story-eval-wide}) shows that, for counterfactual story retelling, post-training is necessary for improving storytelling quality. Since our RRR pipeline uses less than a single epoch from the dataset to reach convergence, it suggests that massive storytelling datasets may not be necessary for post-training. Efforts to curate datasets for d-RLAIF may instead focus on comparatively smaller and query-only datasets. This approach potentially circumvents the ASG bottleneck of limited shared datasets \citep{liu2026narrative}.
    
        Although we apply RRR solely to the counterfactual story retelling task, the method can be speculated to be highly adaptable. By prompting an LLM-as-judge within the d-RLAIF pipeline, the training process allows the student policy model to learn behaviours compatible with the teacher model's conceptual representations. This process requires minimal training data, provided the LLM-as-judge possesses a robust representation of the target behaviour. Using d-RLAIF, methods of effective narrative understanding can provide great assistance in providing the reward signals needed for the training of future ASG models. We believe it to be beneficial to move towards unified knowledge and methods between generation and understanding.  
        
        Future directions can test if adaptions of RRR can generalise to other storytelling tasks (such as open-ended story generation, long-form narrative generation, or interactive storytelling) with different criteria. For instance, models can be trained to adhere to stylistic genre conventions or generate empathetic narratives. In this way, the pipeline itself could potentially serve as a postclassical representation of narrative that models the interaction between text and the reader \citep{piper-etal-2021-narrative}.

    \subsection{Implications for Narrative Theory}
        While we use narrative theory as a foundation for storytelling, the fair-to-moderate agreement empirically highlights that the inherent subjectivity of storytelling may be incompatible with universal benchmarks, and that Todorov's structural theory alone cannot model all narrative qualities (e.g., the ‘Logical’ and ‘Rational’ criteria also draw on cognitive considerations).


        At the same time, the successful RL alignment of LLMs with our narrativity measure to improve task performance suggests that narrativity corresponds to a construct represented within LLMs. Such alignment and performance would be unlikely if narrativity was an arbitrary or structureless construct. This phenomenon merits further investigation by computer scientists and narratologists alike: how a computational model trained on narrative texts comes to exhibit narrative representation.


\section{Related Work}
    \citet{qin-etal-2019-counterfactual}'s TimeTravel dataset and its counterfactual story retelling task have been approached by a number of past works, experimenting with: inference-time optimisations \citep{qin-etal-2020-back, chen2022unsupervised}, multi-step generation \citep{hao2021sketch}, Program-prompting \citep{liu-etal-2023-magic} and ConceptNet \citep{ashwani2024cause}. These past works did not focus on developing a training method which aligns with narrative theory or storytelling principles.

    \citet{wei-etal-2025-igniting} showed that d-RLAIF with GRPO can perform well at improving creative writing for generating Chinese greetings, a task which involves open-ended generation.
        
\section{Conclusion}
    To address the persistent limitations of LLM post-training in Automatic Story Generation, we apply reinforcement learning to teach the policy model how to generate texts that exhibit causal narrative structure. First, we demonstrate that narrativeness in short counterfactual stories can be measured through definable (although not universal or objective) properties--synthesising \citet{todorov19712}’s Theory of Narrative Equilibrium with \citet{prince2003dictionary}’s scalar narrativity into a quantifiable metric based on structural stages (\textbf{RQ1}). Then, we introduce ``Retell Reward Repeat'' (RRR) to align LLMs with these criteria (\textbf{RQ2}) using direct Reinforcement Learning from AI Feedback (d-RLAIF).
    
    Methodologically, our findings suggest that d-RLAIF is a viable post-training paradigm for story generation. Our results across three open-weight 7/8B LLMs show that models post-trained with our narrativity reward signal ($R_N$) consistently outperform SFT and few-shot baselines in output quality, as measured in the counterfactual storytelling task using the criteria: Logical, Rational and Complete\textsubscript{N}. High scores on these criteria show alignment with a blind human preference subset. Future work can adapt RRR to test its generalisability for a broader range of story generation tasks. 
    
     Theoretically, our case study demonstrates both the benefits and limitations of applying \citet{todorov1969grammaire}'s classical structuralist narrative theory to model storytelling, highlighting the need to move beyond textual features. Postclassical narratology expands beyond the limitations of structuralist narratology by emphasising the historical, cultural, and cognitive parameters of storytelling and the reader's role in constructing narratives from linguistic artifacts \citep{meister2011narratology}.

\section*{Limitations}
Our study focuses on counterfactual story reasoning and generation over short five-sentence narratives from the TimeTravel dataset and benchmark \citep{qin-etal-2019-counterfactual}. While this constrained task is suitable for isolating structural causal transformations for a narrative-theory informed RL case study, the generalisability of our findings to other storytelling tasks still needs to be explored.

The human annotation pool was comprised of four paper authors, which introduces potential bias. Model training and dataset curation were conducted by Liu (A), who also contributed to the annotation. Srirag (B), Muston (C), and Joshi (D) were not involved in modelling and strictly contributed to the blind annotation and manuscript drafting.  Prior to annotation, all text orders were randomised and their sources were anonymised.


The study's final model evaluations rely on a proprietary LLM-as-judge (Gemini-3-Flash). While our choice of Gemini as the final evaluator is grounded in an initial benchmarking phase on an annotation subset ($n=200$), where Gemini demonstrated the strongest alignment with human annotators, later comparisons with direct human preference on the final evaluations were conducted on a sample subset of $n=100$ with only one annotator.

We restricted our baseline comparisons to in-context prompting and fine-tuning paradigms, omitting prior TimeTravel systems that optimised primarily for reference-based lexical metrics (e.g., BLEU, ROUGE). Our experiments also exclusively evaluated 7B/8B parameter models; the effectiveness and scaling behaviour of narrative-theoretic reinforcement learning on substantially smaller or larger architectures is left to be explored in future studies.

\section*{Ethics Statement}
   The stories that we study are all in English and refer to predominantly Western contexts; we use a European narrative theory. It is well known in the narrative studies community that cultural differences can affect how we perceive narratives and their underlying themes and structures \citep{aziz2023cross,phillips2025age}. As such, our paper's representation of storytelling does not reflect all cultures.

   Storytelling is a highly effective communication tool that can be used for influence and manipulation. The development of Automatic Story Generation methods, with the goal of matching human storytelling, introduces the risk of harm. This includes: deceptive generation, automated disinformation, and rhetorical manipulation.

   GitHub Copilot and Google AI studio were used to assist with coding tasks and debugging. Microsoft Copilot was used for formatting the LaTeX in the manuscript. Outputs generated by these tools were carefully reviewed and validated by the human authors to maintain correctness.

   We use a publicly available benchmark dataset \citep{qin-etal-2019-counterfactual}, the human annotators are authors on the paper, and we do not have any additional ethical considerations to report. 
   

\section*{Acknowledgements}
This research was supported by the Commonwealth of Australia through an Australian Government Research Training Program Scholarship [DOI: \url{https://doi.org/10.82133/C42F-K220}].

\bibliography{custom}

@inproceedings{qin-etal-2019-counterfactual,
    title = "Counterfactual Story Reasoning and Generation",
    author = "Qin, Lianhui  and
      Bosselut, Antoine  and
      Holtzman, Ari  and
      Bhagavatula, Chandra  and
      Clark, Elizabeth  and
      Choi, Yejin",
    editor = "Inui, Kentaro  and
      Jiang, Jing  and
      Ng, Vincent  and
      Wan, Xiaojun",
    booktitle = "Proceedings of the 2019 Conference on Empirical Methods in Natural Language Processing and the 9th International Joint Conference on Natural Language Processing (EMNLP-IJCNLP)",
    month = nov,
    year = "2019",
    address = "Hong Kong, China",
    publisher = "Association for Computational Linguistics",
    url = "https://aclanthology.org/D19-1509/",
    doi = "10.18653/v1/D19-1509",
    pages = "5043--5053"
}

@inproceedings{chen2022unsupervised,
  title={Unsupervised editing for counterfactual stories},
  author={Chen, Jiangjie and Gan, Chun and Cheng, Sijie and Zhou, Hao and Xiao, Yanghua and Li, Lei},
  booktitle={Proceedings of the AAAI Conference on Artificial Intelligence},
  volume={36},
  number={10},
  pages={10473--10481},
  year={2022}
}

@inproceedings{ashwani2024cause,
  title={Cause and effect: can large language models truly understand causality?},
  author={Ashwani, Swagata and Hegde, Kshiteesh and Mannuru, Nishith Reddy and Sengar, Dushyant Singh and Jindal, Mayank and Kathala, Krishna Chaitanya Rao and Banga, Dishant and Jain, Vinija and Chadha, Aman},
  booktitle={Proceedings of the AAAI Symposium Series},
  volume={4},
  number={1},
  pages={2--9},
  year={2024}
}

@inproceedings{liu-etal-2023-magic,
    title = "The Magic of {IF}: Investigating Causal Reasoning Abilities in Large Language Models of Code",
    author = "Liu, Xiao  and
      Yin, Da  and
      Zhang, Chen  and
      Feng, Yansong  and
      Zhao, Dongyan",
    editor = "Rogers, Anna  and
      Boyd-Graber, Jordan  and
      Okazaki, Naoaki",
    booktitle = "Findings of the Association for Computational Linguistics: ACL 2023",
    month = jul,
    year = "2023",
    address = "Toronto, Canada",
    publisher = "Association for Computational Linguistics",
    url = "https://aclanthology.org/2023.findings-acl.574/",
    doi = "10.18653/v1/2023.findings-acl.574",
    pages = "9009--9022"
}

@inproceedings{qin-etal-2020-back,
    title = "Back to the Future: Unsupervised Backprop-based Decoding for Counterfactual and Abductive Commonsense Reasoning",
    author = "Qin, Lianhui  and
      Shwartz, Vered  and
      West, Peter  and
      Bhagavatula, Chandra  and
      Hwang, Jena D.  and
      Le Bras, Ronan  and
      Bosselut, Antoine  and
      Choi, Yejin",
    editor = "Webber, Bonnie  and
      Cohn, Trevor  and
      He, Yulan  and
      Liu, Yang",
    booktitle = "Proceedings of the 2020 Conference on Empirical Methods in Natural Language Processing (EMNLP)",
    month = nov,
    year = "2020",
    address = "Online",
    publisher = "Association for Computational Linguistics",
    url = "https://aclanthology.org/2020.emnlp-main.58/",
    doi = "10.18653/v1/2020.emnlp-main.58",
    pages = "794--805"
}

@inproceedings{mostafazadeh-etal-2016-corpus,
    title = "A Corpus and Cloze Evaluation for Deeper Understanding of Commonsense Stories",
    author = "Mostafazadeh, Nasrin  and
      Chambers, Nathanael  and
      He, Xiaodong  and
      Parikh, Devi  and
      Batra, Dhruv  and
      Vanderwende, Lucy  and
      Kohli, Pushmeet  and
      Allen, James",
    editor = "Knight, Kevin  and
      Nenkova, Ani  and
      Rambow, Owen",
    booktitle = "Proceedings of the 2016 Conference of the North {A}merican Chapter of the Association for Computational Linguistics: Human Language Technologies",
    month = jun,
    year = "2016",
    address = "San Diego, California",
    publisher = "Association for Computational Linguistics",
    url = "https://aclanthology.org/N16-1098/",
    doi = "10.18653/v1/N16-1098",
    pages = "839--849"
}

@inproceedings{lee2024rlaif,
  title={RLAIF vs. RLHF: scaling reinforcement learning from human feedback with AI feedback},
  author={Lee, Harrison and Phatale, Samrat and Mansoor, Hassan and Mesnard, Thomas and Ferret, Johan and Lu, Kellie and Bishop, Colton and Hall, Ethan and Carbune, Victor and Rastogi, Abhinav and others},
  booktitle={Proceedings of the 41st International Conference on Machine Learning},
  pages={26874--26901},
  year={2024}
}

@inproceedings{wei-etal-2025-igniting,
    title = "Igniting Creative Writing in Small Language Models: {LLM}-as-a-Judge versus Multi-Agent Refined Rewards",
    author = "Wei, Xiaolong  and
      Lu, Bo  and
      Zhang, Xingyu  and
      Zhao, Zhejun  and
      Shen, Dongdong  and
      Xia, Long  and
      Yin, Dawei",
    editor = "Christodoulopoulos, Christos  and
      Chakraborty, Tanmoy  and
      Rose, Carolyn  and
      Peng, Violet",
    booktitle = "Proceedings of the 2025 Conference on Empirical Methods in Natural Language Processing",
    month = nov,
    year = "2025",
    address = "Suzhou, China",
    publisher = "Association for Computational Linguistics",
    url = "https://aclanthology.org/2025.emnlp-main.868/",
    doi = "10.18653/v1/2025.emnlp-main.868",
    pages = "17171--17197",
    ISBN = "979-8-89176-332-6"
}

@article{todorov19712,
  title={The 2 principles of narrative},
  author={Todorov, Tzvetan},
  journal={diacritics},
  pages={37--44},
  year={1971},
  publisher={JSTOR}
}

@inproceedings{chakrabarty2024art,
  title={Art or artifice? large language models and the false promise of creativity},
  author={Chakrabarty, Tuhin and Laban, Philippe and Agarwal, Divyansh and Muresan, Smaranda and Wu, Chien-Sheng},
  booktitle={Proceedings of the 2024 CHI Conference on Human Factors in Computing Systems},
  pages={1--34},
  year={2024}
}

@article{zec2017high,
  title={High agreement and high prevalence: the paradox of Cohen’s kappa},
  author={Zec, Slavica and Soriani, Nicola and Comoretto, Rosanna and Baldi, Ileana},
  journal={The open nursing journal},
  volume={11},
  pages={211},
  year={2017}
}

@book{gwet2014handbook,
  title={Handbook of inter-rater reliability: The definitive guide to measuring the extent of agreement among raters},
  author={Gwet, Kilem L},
  year={2014},
  publisher={Advanced Analytics, LLC}
}

@article{alexandru2025atla,
  title={Atla selene mini: A general purpose evaluation model},
  author={Alexandru, Andrei and Calvi, Antonia and Broomfield, Henry and Golden, Jackson and Dai, Kyle and Leys, Mathias and Burger, Maurice and Bartolo, Max and Engeler, Roman and Pisupati, Sashank and others},
  journal={arXiv preprint arXiv:2501.17195},
  year={2025}
}

@article{pombal2025m,
  title={M-Prometheus: A Suite of Open Multilingual LLM Judges},
  author={Pombal, Jos{\'e} and Yoon, Dongkeun and Fernandes, Patrick and Wu, Ian and Kim, Seungone and Rei, Ricardo and Neubig, Graham and Martins, Andr{\'e} FT},
  journal={arXiv preprint arXiv:2504.04953},
  year={2025}
}

@misc{google2025gemini3flash,
  author       = {Google},
  title        = {{Introducing Gemini 3 Flash: Benchmarks, Global Availability}},
  howpublished = {\url{https://blog.google/products/gemini/gemini-3-flash/}},
  month        = dec,
  year         = {2025},
  note         = {Accessed 2025-12-27}
}

@article{olmo2025olmo,
  title={Olmo 3},
  author={Ettinger, Allyson and Bertsch, Amanda and Kuehl, Bailey and Graham, David and Heineman, David and Groeneveld, Dirk and Brahman, Faeze and Timbers, Finbarr and Ivison, Hamish and others},
  journal={arXiv preprint arXiv:2512.13961},
  year={2025}
}

@article{yang2025qwen3,
  title={Qwen3 technical report},
  author={Yang, An and Li, Anfeng and Yang, Baosong and Zhang, Beichen and Hui, Binyuan and Zheng, Bo and Yu, Bowen and Gao, Chang and Huang, Chengen and Lv, Chenxu and others},
  journal={arXiv preprint arXiv:2505.09388},
  year={2025}
}

@article{grattafiori2024llama,
  title={The llama 3 herd of models},
  author={Grattafiori, Aaron and Dubey, Abhimanyu and Jauhri, Abhinav and Pandey, Abhinav and Kadian, Abhishek and Al-Dahle, Ahmad and Letman, Aiesha and Mathur, Akhil and Schelten, Alan and Vaughan, Alex and others},
  journal={arXiv preprint arXiv:2407.21783},
  year={2024}
}

@article{shao2024deepseekmath,
  title={Deepseekmath: Pushing the limits of mathematical reasoning in open language models},
  author={Shao, Zhihong and Wang, Peiyi and Zhu, Qihao and Xu, Runxin and Song, Junxiao and Bi, Xiao and Zhang, Haowei and Zhang, Mingchuan and Li, YK and Wu, Yang and others},
  journal={arXiv preprint arXiv:2402.03300},
  year={2024}
}

@article{hu2022lora,
  title={Lora: Low-rank adaptation of large language models.},
  author={Hu, Edward J and Shen, Yelong and Wallis, Phillip and Allen-Zhu, Zeyuan and Li, Yuanzhi and Wang, Shean and Wang, Lu and Chen, Weizhu and others},
  journal={ICLR},
  volume={1},
  number={2},
  pages={3},
  year={2022}
}

@inproceedings{li2013story,
  title={Story generation with crowdsourced plot graphs},
  author={Li, Boyang and Lee-Urban, Stephen and Johnston, George and Riedl, Mark},
  booktitle={Proceedings of the AAAI Conference on Artificial Intelligence},
  volume={27},
  number={1},
  pages={598--604},
  year={2013}
}

@inproceedings{He2023DeBERTaV3ICLR,
  title     = {DeBERTaV3: Improving DeBERTa using ELECTRA-Style Pre-Training with Gradient-Disentangled Embedding Sharing},
  author    = {He, Pengcheng and Gao, Jianfeng and Chen, Weizhu},
  booktitle = {International Conference on Learning Representations (ICLR)},
  year      = {2023},
  url       = {https://openreview.net/forum?id=sE7-XhLxHA}
}

@misc{meister2011narratology,
  author       = {Jan Christoph Meister},
  title        = {Narratology},
  howpublished = {The Living Handbook of Narratology},
  year         = {2011},
  note         = {rev. 19 January 2014},
  url          = {https://www-archiv.fdm.uni-hamburg.de/lhn/node/48.html},
  urldate      = {2026-01-03}
}

@article{hineline2018narrative,
  title={Narrative: Why it’s important, and how it works},
  author={Hineline, Philip N},
  journal={Perspectives on Behavior Science},
  volume={41},
  number={2},
  pages={471--501},
  year={2018},
  publisher={Springer}
}

@article{aziz2023cross,
  title={Cross-cultural narratology: A comparative study of storytelling techniques in Eastern and Western literature},
  author={Aziz, Sardar Khawar},
  journal={Journal of Asian Development Studies},
  volume={12},
  number={4},
  pages={742--753},
  year={2023}
}

@article{phillips2025age,
  title={Age and cultural differences in the relationship between reading and theory of mind},
  author={Phillips, Louise H and Lawrie, Louisa and Suchomelova, Zuzana and Hein{\"a}maa, Sara and O'Dwyer, Amy and Yong, Min Hooi},
  journal={Poetics},
  volume={109},
  pages={101984},
  year={2025},
  publisher={Elsevier}
}

@inproceedings{papineni2002bleu,
  title        = {Bleu: A Method for Automatic Evaluation of Machine Translation},
  author       = {Papineni, Kishore and Roukos, Salim and Ward, Todd and Zhu, Wei‑Jing},
  booktitle    = {Proceedings of the 40th Annual Meeting of the Association for Computational Linguistics},
  year         = {2002},
  pages        = {311--318},
  address      = {Philadelphia, Pennsylvania, USA},
  publisher    = {Association for Computational Linguistics},
  doi          = {10.3115/1073083.1073135},
  url          = {https://aclanthology.org/P02-1040/}
}

@inproceedings{lin2004rouge,
  title        = {ROUGE: A Package for Automatic Evaluation of Summaries},
  author       = {Lin, Chin‑Yew},
  booktitle    = {Text Summarization Branches Out (Workshop at ACL)},
  year         = {2004},
  pages        = {74--81},
  address      = {Barcelona, Spain},
  publisher    = {Association for Computational Linguistics},
  url          = {https://aclanthology.org/W04-1013/}
}

@inproceedings{ZhangKWWA20,
  title     = {BERTScore: Evaluating Text Generation with BERT},
  author    = {Zhang, Tianyi and Kishore, Varsha and Wu, Felix and Weinberger, Kilian Q. and Artzi, Yoav},
  booktitle = {International Conference on Learning Representations (ICLR) 2020},
  year      = {2020},
  publisher = {OpenReview.net},
  url       = {https://openreview.net/forum?id=SkeHuCVFDr}
}

@inproceedings{lal2024cat,
  title={CaT-bench: Benchmarking language model understanding of causal and temporal dependencies in plans},
  author={Lal, Yash Kumar and Cohen, Vanya and Chambers, Nathanael and Balasubramanian, Niranjan and Mooney, Ray},
  booktitle={Proceedings of the 2024 Conference on Empirical Methods in Natural Language Processing},
  pages={19336--19354},
  year={2024}
}

@inproceedings{lampinen-etal-2022-language,
    title = "Can language models learn from explanations in context?",
    author = "Lampinen, Andrew  and
      Dasgupta, Ishita  and
      Chan, Stephanie  and
      Mathewson, Kory  and
      Tessler, Mh  and
      Creswell, Antonia  and
      McClelland, James  and
      Wang, Jane  and
      Hill, Felix",
    editor = "Goldberg, Yoav  and
      Kozareva, Zornitsa  and
      Zhang, Yue",
    booktitle = "Findings of the Association for Computational Linguistics: EMNLP 2022",
    month = dec,
    year = "2022",
    address = "Abu Dhabi, United Arab Emirates",
    publisher = "Association for Computational Linguistics",
    url = "https://aclanthology.org/2022.findings-emnlp.38/",
    doi = "10.18653/v1/2022.findings-emnlp.38",
    pages = "537--563"
}

@article{barthes1967death,
  author  = {Barthes, Roland},
  title   = {The Death of the Author},
  journal = {Aspen},
  year    = {1967},
  number  = {5--6},
  translator = {Howard, Richard}
}

@inproceedings{
    sui2026llms,
    title={{LLM}s Exhibit Significantly Lower Uncertainty in Creative Writing Than Professional Writers},
    author={Peiqi Sui},
    booktitle={Third Conference on Language Modeling},
    year={2026},
    url={https://openreview.net/forum?id=n1IFqJ0tFb}
}

@inproceedings{chakrabarty2025can,
  title={Can ai writing be salvaged? mitigating idiosyncrasies and improving human-ai alignment in the writing process through edits},
  author={Chakrabarty, Tuhin and Laban, Philippe and Wu, Chien-Sheng},
  booktitle={Proceedings of the 2025 CHI Conference on Human Factors in Computing Systems},
  pages={1--33},
  year={2025}
}

@article{atmakuru2024cs4,
  title={Cs4: Measuring the creativity of large language models automatically by controlling the number of story-writing constraints},
  author={Atmakuru, Anirudh and Nainani, Jatin and Bheemreddy, Rohith Siddhartha Reddy and Lakkaraju, Anirudh and Yao, Zonghai and Zamani, Hamed and Chang, Haw-Shiuan},
  journal={arXiv preprint arXiv:2410.04197},
  year={2024}
}

@inproceedings{chhun2022human,
  title={Of human criteria and automatic metrics: A benchmark of the evaluation of story generation},
  author={Chhun, Cyril and Colombo, Pierre and Suchanek, Fabian and Clavel, Chlo{\'e}},
  booktitle={Proceedings of the 29th International Conference on Computational Linguistics},
  pages={5794--5836},
  year={2022}
}

@article{liu2026narrative,
  title={Narrative Theory-Driven LLM Methods for Automatic Story Generation and Understanding: A Survey},
  author={Liu, David Y and Joshi, Aditya and Dawson, Paul},
  journal={arXiv preprint arXiv:2602.15851},
  year={2026}
}

@article{netisopakul2023comparison,
  title={Comparison of evaluation metrics for short story generation},
  author={Netisopakul, Ponrudee and Taoto, Usanisa},
  journal={IEEE access},
  volume={11},
  pages={140253--140269},
  year={2023},
  publisher={IEEE}
}

@inproceedings{marco2025reader,
  title={The reader is the metric: How textual features and reader profiles explain conflicting evaluations of AI creative writing},
  author={Marco, Guillermo and Gonzalo, Julio and Fresno, V{\'\i}ctor},
  booktitle={Findings of the Association for Computational Linguistics: ACL 2025},
  pages={25432--25449},
  year={2025}
}

@inproceedings{
    lu2025ai,
    title={{AI} as Humanity{\textquoteright}s Salieri: Quantifying Linguistic Creativity of Language Models via Systematic Attribution of Machine Text against Web Text},
    author={Ximing Lu and Melanie Sclar and Skyler Hallinan and Niloofar Mireshghallah and Jiacheng Liu and Seungju Han and Allyson Ettinger and Liwei Jiang and Khyathi Chandu and Nouha Dziri and Yejin Choi},
    booktitle={The Thirteenth International Conference on Learning Representations},
    year={2025},
    url={https://openreview.net/forum?id=ilOEOIqolQ}
}

@inproceedings{hao2021sketch,
  title={Sketch and customize: A counterfactual story generator},
  author={Hao, Changying and Pang, Liang and Lan, Yanyan and Wang, Yan and Guo, Jiafeng and Cheng, Xueqi},
  booktitle={Proceedings of the AAAI conference on artificial intelligence},
  volume={35},
  number={14},
  pages={12955--12962},
  year={2021}
}

@book{prince2003dictionary,
  title={A dictionary of narratology},
  author={Prince, Gerald},
  year={2003},
  publisher={U of Nebraska Press}
}

@book{todorov1969grammaire,
  title={Grammaire du D{\'e}cam{\'e}ron},
  author={Todorov, T.},
  isbn={9783110981346},
  lccn={71091208},
  series={Approaches to semiotics},
  url={https://books.google.com.au/books?id=YyddAAAAMAAJ},
  year={1969},
  publisher={Mouton}
}

@article{liu2025eliciting,
  title={Eliciting and improving the causal reasoning abilities of large language models with conditional statements},
  author={Liu, Xiao and Yin, Da and Zhang, Chen and Zhao, Dongyan and Feng, Yansong},
  journal={Computational Linguistics},
  volume={51},
  pages={467--504},
  year={2025}
}

@inproceedings{piper-etal-2021-narrative,
    title = "Narrative Theory for Computational Narrative Understanding",
    author = "Piper, Andrew  and
      So, Richard Jean  and
      Bamman, David",
    editor = "Moens, Marie-Francine  and
      Huang, Xuanjing  and
      Specia, Lucia  and
      Yih, Scott Wen-tau",
    booktitle = "Proceedings of the 2021 Conference on Empirical Methods in Natural Language Processing",
    month = nov,
    year = "2021",
    address = "Online and Punta Cana, Dominican Republic",
    publisher = "Association for Computational Linguistics",
    url = "https://aclanthology.org/2021.emnlp-main.26/",
    doi = "10.18653/v1/2021.emnlp-main.26",
    pages = "298--311"
}

\appendix

\section{License}
    The TimeTravel dataset is used under the MIT License.
    Hugging face's Transformers and TRL libraries are used under the Apache License 2.0 license.
    Llama 3.1 is used under the LLaMA 3.1 Community License.
    Olmo 3, Qwen 3 and Selene-1-mini are used under the Apache License 2.0.

    We use these artifacts in a way that's consistent with intended use for the purpose of academic research.

\section{TimeTravel Dataset}
\label{sec:TimeTravel}

    The TimeTravel dataset is constructed from the ROCStories corpus \citep{mostafazadeh-etal-2016-corpus} which consists of over 100k human written five-sentence short stories. In TimeTravel, there are 1.87k items in the test set, 1.87k items in the validation set, and 16.8k items in the supervised training set, all with human written counterfactual endings. Furthermore, the entire 100k ROC stories can be used for unsupervised training.
    
    The dataset includes narratives that may allude to violent themes, and some generated outputs could be distressing. To the best of our knowledge, these stories are entirely fictional and do not represent real individuals.

\section{Computational Budget}

    Inference and d-RLAIF and SFT training was conducted on a single H200 GPU. The duration of each training process was between 30 minutes and 4 hours. 

    Evaluation using Gemini-3-Flash was done using OpenRouter API, which cost roughly \$3 per testset (n = 1871).

\section{Hyperparameters}

    We detail the hyperparameters applied for supervised fine-tuning (SFT) and GRPO-based deep reinforcement learning from AI feedback (d-RLAIF). All training and inference were performed using Hugging Face’s Transformers and TRL libraries. The same hyperparameter configuration was maintained across all experimental settings. Early stopping was implemented for both SFT and d-RLAIF.

\subsection{Supervised Fine-Tuning (SFT)}

    For SFT, we train the model using the AdamW optimiser with a fixed learning rate and Low-Rank Adaptation (LoRA) parameterisation \cite{hu2022lora}. Training is conducted for a single fine-tuning stage, with evaluation performed periodically on a held-out validation set. The best model checkpoint is selected based on validation loss using early stopping.
    
    LoRA-specific hyperparameters are:
    \begin{itemize}[style=unboxed,leftmargin=0cm,labelindent=0cm,itemsep=0cm,parsep=0cm] 
        \item Rank $r = 64$
        \item Alpha $\alpha = 128$
        \item Dropout $= 0.05$
    \end{itemize}
    
    Other training hyperparameters are:
    \begin{itemize}[style=unboxed,leftmargin=0cm,labelindent=0cm,itemsep=0cm,parsep=0cm] 
        \item Per-device batch size $= 8$
        \item Gradient accumulation steps $= 8$
        \item Learning rate $= 1\times 10^{-4}$
        \item Number of epochs $= 1$
        \item Maximum sequence length $= 640$ tokens
    \end{itemize}
    
    Mixed-precision training is enabled using bfloat16 (bf16). Gradient check pointing is used to reduce memory usage. No extensive hyperparameter tuning was performed; values were chosen heuristically.
    
    \subsection{d-RLAIF with GRPO}
    
    For d-RLAIF, we employ GRPO with LoRA parameterisation for the policy model. At each training step, 16 candidate completions are generated per prompt, and early stopping is used based on evaluation metrics.
    
    LoRA-specific hyperparameters for d-RLAIF:
    \begin{itemize}[style=unboxed,leftmargin=0cm,labelindent=0cm,itemsep=0cm,parsep=0cm] 
        \item Rank $r = 64$
        \item Alpha $\alpha = 128$
        \item Dropout $= 0.05$
    \end{itemize}
    
    Other training hyperparameters are:
    \begin{itemize}[style=unboxed,leftmargin=0cm,labelindent=0cm,itemsep=0cm,parsep=0cm] 
        \item Per-device batch size $= 24$
        \item Gradient accumulation steps $= 2$
        \item Learning rate $= 5\times 10^{-6}$
        \item Number of epochs $= 1$
        \item Maximum prompt and completion length $= 512$ tokens
    \end{itemize}
    
    Mixed-precision training using bfloat16 (bf16) was enabled. No extensive hyperparameter search was performed; hyperparameters were selected heuristically. All experiments use fixed hyperparameters across conditions. Sensitivity to hyperparameter choices is left to future work.

\section{Annotator Guidelines}
\label{sec:annotator_instructions}
    To ensure consistent understanding of definitions, annotators are provided with instructions and examples before continuing into the main workflow, which consist of two primary phases: (1) Reading and Tagging, and (2) Rating.
    
    \subsection{Narrative State Tagging}
        Annotators are instructed to read both the original and rewritten stories, highlighting substrings according to Todorov's Narrative Theory. They are provided with the following criteria for the five narrative stages:
    
    \begin{description}[style=unboxed,leftmargin=0cm,labelindent=0cm,itemsep=0cm,parsep=0cm]
        \item[Equilibrium:] An initial state before any transformation that is then changed by a disruption.
        \item[Disruption of Equilibrium:] The event/condition that disrupts equilibrium and initiates transformations.
        \item[Recognition of the Disruption:] A character or narrator recognises that the disruption has occurred.
        \item[Attempt to Resolve Disruption:] Actions taken in an attempt to address or resolve the disruption.
        \item[Return to New Equilibrium:] A new equilibrium state reached after the transformation.
    \end{description}
    
    \paragraph{Key Tagging Guidelines}
    \begin{description}[style=unboxed,leftmargin=0cm,labelindent=0cm,itemsep=0cm,parsep=0cm]
        \item[1:] Tagging every single word or sentence is not required; only highlight parts specifically relevant to that stage.
        \item[2:] Some stages may be implicit and cannot be tagged.
        \item[3:] Filler or background detail can be left untagged.
    \end{description}

    \subsection{Common Variations and Visual Examples}
    \label{subsec:variations}
    We provide annotators with annotated examples of different narrative structural variations.
    
    \paragraph{Variation 1: In Medias Res}
    The narrative begins with the Disruption of Equilibrium, skipping the initial equilibrium. \\
    \textit{Example:} ``\hldis{The kitchen pipe burst, flooding the tiles in seconds.} \hlrec{I saw the water reaching the carpet and knew the basement was next.} \hlatt{I grabbed a wrench and tightened the main valve.} \hlnew{Within an hour, the plumber arrived and the house was dry again.}''
    
    \paragraph{Variation 2: Non-Linear Narrative}
    Events are revealed out of order, often starting with the resolution or final effort. \\
    \textit{Example:} ``\hlatt{I crossed the finish line and collapsed onto the grass.} \hleq{Years of regular jogs had kept me healthy} \hldis{until a sudden ankle sprain last winter.} \hlrec{Sitting in the doctor's office back then, I decided I would run a marathon to prove I could heal.} \hlnew{Today, my medal finally felt real.}''

    \paragraph{Variation 3: Cliffhanger Endings}
    The text ends abruptly during an Attempt to Resolve Disruption, with no resolution shown. \\
    \textit{Example:} ``\hleq{The exam room was quiet as the students wrote.} \hldis{Suddenly, the fire alarm screeched throughout the hall.} \hlrec{The proctor stood up and realized there was smoke beneath the door.} \hlatt{She grabbed the evacuation log and shouted for everyone to follow her into the stairwell...}''
    
    \subsection{Rating/Evaluation Guide}
    After tagging, annotators evaluate the edited text based on its adherence to specific criteria. They are instructed to focus strictly on whether the criteria are met, not on the stylistic quality of the writing. 

    \paragraph{Logical:}

    The text must contain no internal contradictions. For example:
    
    \begin{tcolorbox}[colback=gray!5, colframe=gray!40, boxrule=0.4pt,
                      left=6pt,right=6pt,top=4pt,bottom=4pt,enhanced,
                      sharp corners]
        \textbf{Logical:} “Alex lost her job, but since she's an optimist, she walked home smiling.”
    
        \medskip
    
        \textbf{Not Logical:} “Alex lost the job that meant everything to her. She walked home with a smile.” (Illogical without context.)
    \end{tcolorbox}

    \paragraph{Rational:}

    The text must follow a coherent Todorovian progression, with no contextual or causal disconnection. All events should be interpretable as belonging to compatible narrative stages. For example:

    \begin{tcolorbox}[colback=gray!5, colframe=gray!40, boxrule=0.4pt,
                      left=6pt,right=6pt,top=4pt,bottom=4pt,enhanced,
                      sharp corners]
        \textbf{Rational:} “The oven broke. I ordered pizza. We had a great meal.” (Clear causal and contextual continuity.)

        \medskip

        \textbf{Not Rational:} “I was eating dinner. The waiter spilt wine. I finally got my car fixed.” (Narrative jumps without coherent stage progression.)
    \end{tcolorbox}

    \paragraph{Complete\textsubscript{N}:}

    The text must preserve all key Todorovian stage(s) from the original narrative. An edited text is complete only if it reproduces all essential stages, even if the specific events differ. For example:

    \begin{tcolorbox}[colback=gray!5, colframe=gray!40, boxrule=0.4pt,
                      left=6pt,right=6pt,top=4pt,bottom=4pt,enhanced,
                      sharp corners]
        \textbf{Complete:} Orig: \hleq{Health} $\rightarrow$ \hldis{Flu} $\rightarrow$ \hlnew{Recovery}.  
        Edit: \hleq{Health} $\rightarrow$ \hldis{Accident} $\rightarrow$ \hlatt{Surgery} $\rightarrow$ \hlnew{Death}. (All original stages preserved.)

        \medskip

        \textbf{Incomplete:} Orig: \hleq{Health} $\rightarrow$ \hldis{Flu} $\rightarrow$ \hlnew{Recovery}.  
        Edit: \hleq{Health} $\rightarrow$ \hldis{Accident} $\rightarrow$ \hlatt{Surgery}. (Missing the New Equilibrium stage.)
    \end{tcolorbox}

    \subsection{Handling Ambiguity}
    Annotators were given an \textbf{Unsure} (rating=2) option, to be used when the text is ambiguous. 

    \begin{table}[h]
      \centering
      \begin{tabularx}{\linewidth}{@{}X X
          S[table-format=2.1]
          S[table-format=2.1]
          S[table-format=2.1]@{}}
        \toprule
        \textbf{Metric} & \textbf{Rating} &
        {\textbf{A (\%)}} & {\textbf{B (\%)}} & {\textbf{C (\%)}} \\
        \midrule
    
        \multirow{3}{*}{Logical}
          & Agree    & 85.5 & 72.0 & 95.5 \\
          & Neutral  &  7.0 &  1.0 &  0.0 \\
          & Disagree &  7.5 & 27.0 &  4.5 \\
        \cmidrule(lr){1-5}
    
        \multirow{3}{*}{Rational}
          & Agree    & 60.5 & 72.0 & 67.5 \\
          & Neutral  & 19.5 &  6.0 &  0.0 \\
          & Disagree & 20.0 & 22.0 & 32.5 \\
        \cmidrule(lr){1-5}
    
        \multirow{3}{*}{Complete}
          & Agree    & 56.0 & 64.0 & 49.0 \\
          & Neutral  &  7.5 &  2.5 &  0.0 \\
          & Disagree & 36.5 & 33.5 & 51.0 \\
        \cmidrule(lr){1-5}
    
        \multirow{3}{*}{min\textsubscript{LRC}}
          & Agree    & 36.5 & 47.0 & 32.5 \\
          & Neutral  & 18.0 &  2.5 &  0.0 \\
          & Disagree & 45.5 & 50.5 & 67.5 \\
        \cmidrule(lr){1-5}
    
        \multirow{5}{*}{Narrativity}
          & Agree+    & 49.0 & 53.5 & 14.0 \\
          & Agree     & 31.5 & 27.5 & 42.0 \\
          & Neutral   &  8.0 &  9.0 & 25.5 \\
          & Disagree  &  0.0 &  1.0 &  2.5 \\
          & Disagree- & 11.5 &  9.0 & 16.0 \\
        \bottomrule
      \end{tabularx}
    
      \caption{Distribution: 3 human annotators (A, B, C), percentages shown (200 per metric per annotator). Agree+ is Strongly Agree; Disagree- is Strongly Disagree.}
      \label{tab:human_distribution_3}
    \end{table}

    \begin{table*}[h]
        \centering
        \begin{tabular}{lccc}
        \hline
        \textbf{Criteria} & \textbf{AC2} & \textbf{\% Agree} & \textbf{$\kappa$} \\
        \hline
        Logical        & 0.7648 [0.71, 0.82] & 0.7967 & 0.1797 [0.0567, 0.2968] \\
        Rational       & 0.7082 [0.65, 0.76] & 0.7792 & 0.4068 [0.3024, 0.5035] \\
        Complete\textsubscript{N} 
                       & 0.5611 [0.50, 0.62] & 0.6750 & 0.3180 [0.2322, 0.4129] \\
        min\textsubscript{LRC}
                       & 0.5544 [0.49, 0.61] & 0.6767 & 0.2981 [0.2116, 0.3851] \\
        Narrativity    & 0.7415 [0.69, 0.79] & 0.8660 & 0.3764 [0.2690, 0.4796] \\
        \hline
        \end{tabular}
        \caption{Inter-annotator agreement measured by Gwet's AC2, \% Agreement, and Quadratic Weighted $\kappa$. Values in brackets represent the 95\% Confidence Interval (CI).}
        \label{tab:human_annotation}
    \end{table*}

    \begin{table*}[h]
        \centering
        \small 
        \begin{tabular}{lrrrrrr}
        \toprule
        Annotator & Logical & Rational & Complete\textsubscript{N} & Overall & min\textsubscript{LRC} & Narrativity \\
        \midrule
        Humans Only
        & 0.76/0.18 
        & 0.71/0.41 
        & 0.56/0.32 
        &  
        & 0.55/0.30 
        & 0.74/0.38 \\
    
        \midrule
    
        w/ Selene-1-mini
        & 0.80/0.09 
        & 0.71/0.27 
        & 0.58/0.25 
        & 0.49/0.20 
        & 0.54/0.20 
        & 0.73/0.31 \\
    
        w/ M-Prometheus
        & 0.81/0.12 
        & 0.71/0.30 
        & 0.56/0.25 
        & 0.47/0.21 
        & 0.48/0.23 
        & 0.73/0.28 \\
    
        w/ Gemini-3-Flash
        & 0.78/0.22 
        & 0.71/0.36 
        & 0.59/0.36 
        & 0.53/0.29 
        & 0.55/0.31 
        & 0.77/0.40 \\
        
        \bottomrule
        \end{tabular}
        \caption{Inter-annotator agreement, reported by \textit{AC2/$\kappa$}. “w/” indicates when an LLM-evaluator is added.}
        \label{tab:llm_as_judge_vs_human}
    \end{table*}

    \begin{table*}[h]
        \centering
        \small 
        \setlength{\tabcolsep}{4pt} 
        \begin{tabular}{lccccc}
        \toprule
        \textbf{Author} & \textbf{Logical} & \textbf{Rational} & \textbf{Complete$_N$} & \textbf{min$_{LRC}$} & \textbf{Narrativity} \\
        \midrule
        human        & 2.98 / 2.54 / 2.96 & 2.72 / 2.66 / 2.52 & 2.82 / 2.72 / 2.28 & 2.58 / 2.30 / 1.96 & 4.70 / 4.62 / 3.72 \\
        Olmo        & 2.80 / 2.50 / 2.96 & 2.52 / 2.66 / 2.48 & 1.92 / 2.04 / 2.00 & 1.70 / 1.84 / 1.72 & 3.84 / 4.00 / 3.36 \\
        Qwen        & 2.64 / 2.40 / 2.92 & 2.36 / 2.52 / 2.36 & 2.50 / 2.54 / 2.12 & 2.12 / 2.12 / 1.68 & 4.22 / 4.32 / 3.50 \\
        Llama       & 2.70 / 2.36 / 2.80 & 2.02 / 2.16 / 2.04 & 1.54 / 1.92 / 1.52 & 1.24 / 1.60 / 1.24 & 3.50 / 3.68 / 2.84 \\
        \bottomrule
        \end{tabular}
        \caption{Human and LLM author comparison across evaluation dimensions with 3 human annotators (A / B / C).}
        \label{tab:ranking_200}
    \end{table*}

    \begin{table*}[h]
        \centering
        \small 
        \setlength{\tabcolsep}{4pt} 
        \begin{tabular}{lcccccc}
        \toprule
        \textbf{Model} & \textbf{Logical} & \textbf{Rational} & \textbf{Complete$_N$} & \textbf{Overall} & \textbf{min$_{LRC}$} & \textbf{Narrativity} \\
        \midrule
        human    & 2.80 / 2.94 / 3.00 & 2.92 / 2.94 / 2.96 & 2.96 / 2.38 / 2.82 & 2.52 / 2.46 / 2.62 & 2.72 / 2.38 / 2.80 & 4.90 / 3.62 / 5.00 \\
        Qwen    & 2.60 / 3.00 / 2.96 & 2.80 / 3.00 / 3.00 & 2.36 / 2.54 / 2.86 & 2.40 / 2.74 / 2.76 & 2.04 / 2.54 / 2.82 & 4.78 / 3.42 / 4.88 \\
        Olmo    & 2.72 / 3.00 / 3.00 & 2.80 / 2.98 / 3.00 & 1.88 / 2.14 / 2.72 & 2.20 / 2.40 / 2.76 & 1.76 / 2.14 / 2.72 & 4.38 / 3.26 / 4.76 \\
        Llama & 2.68 / 2.92 / 2.94 & 2.24 / 2.72 / 2.52 & 1.52 / 1.90 / 2.22 & 1.60 / 2.10 / 2.28 & 1.36 / 1.86 / 2.12 & 3.82 / 2.76 / 4.18 \\
        \bottomrule
        \end{tabular}
        \caption{Model comparison across evaluation dimensions with LLM judges (Gemini / Selene / Prometheus).}
        \label{tab:v9_rank}
    \end{table*}

    \subsection{Additional Agreement Statistics}

    Since the collection of human-authored and LLM-generated stories is logical and rational in most cases, the rating distributions are skewed (e.g., the average rating for Logical is 2.7/3 while the average for narrativity is 3.9/5. See Appendix Table \ref{tab:human_distribution_3}), leading to the well-known prevalence/bias paradox \cite{zec2017high} for $\kappa$ (e.g. $\kappa=0.18$ despite $80\%$ agreement, see Appendix Table \ref{tab:human_annotation}). 
    
    We want the dataset to test whether LLM-evaluators can distinguish between genuine human-level narratives and modern LLM generations. To accurately reflect the expected distribution of scores that the LLM-evaluator will see during d-RLAIF and evaluation, we do not alter the distributions and instead report \citet{gwet2014handbook}'s AC2 alongside $\kappa$ for inter-annotator reliability (top row of Appendix Table \ref{tab:llm_as_judge_vs_human}). 
    
    Unlike $\kappa$, which mathematically conflates high class prevalence with a high probability of chance agreement, AC2 estimates chance agreement under the assumption that raters do not guess randomly when they are certain of a prevalent class, thereby providing a useful complementary metric for skewed distributions.

    Inter-annotator agreement across the evaluation criteria ranged from fair to moderate (Appendix Table \ref{tab:llm_as_judge_vs_human}), with AC2 scores between 0.55–0.76 and $\kappa$ values between 0.18–0.41 (or 0.30-0.41 if we treat the Logical criterion as an outlier since the distribution is skewed, leading to a low $\kappa$). 

\section{Few-Shot Method}
\label{appendix-few-shot}

    To select examples for our few-shot method, we use TimeTravel's supervised training set. For each example in the training set, we use DeBERTa-v3-small \citep{He2023DeBERTaV3ICLR} to calculate the similarity between the retelling and the original (using BERTScore \cite{ZhangKWWA20}). 
    
    We then rank the examples from most different to least different. We select the 1st, 500th and 1000th examples to diversify our 3 shot prompt to include examples that require high, medium, and low changes to the original ending.

\section{LLM-as-judge}  
    In addition to Spearman's $\rho$ (Section \ref{sec:llm_as_judge}), we also evaluate the models' alignment based on AC2 \& $\kappa$, measured by examining the agreement scores when the LLM-as-judge is included in a 4-way calculation with the 3 human annotators (Table \ref{tab:llm_as_judge_vs_human}). 
    
    We observe that Gemini is the only model which is able to consistently improve alignment measured by both metrics in Complete\textsubscript{N}, Overall, min\textsubscript{LRC} and Narrativity. 
    
    Additionally, when we again categorise the ratings by the author, we see that only Gemini rates human authored stories with the highest scores, while Selene and Prometheus only rank human authored stories first in narrativity (Appendix Table \ref{tab:v9_rank}). We choose to save Gemini for final evaluation, given the evidence that it is our most reliable and aligned LLM-as-judge.
    
    While we ultimately only use a `no-reasoning' prompt for the LLM-as-judge for training and evaluation, we initially tested two different output formats: 

    \begin{enumerate}[style=unboxed,leftmargin=0cm,labelindent=0cm,itemsep=0cm,parsep=0cm]
        \item Generate the reasoning and then the score
        \item Generate the score only\footnote{We instruct the LLM to output the score then the reasoning because post-hoc reasoning can improve accuracy \cite{lampinen-etal-2022-language,lal2024cat}. We stop inference after the score is generated, since it is impossible for the subsequent tokens to change the already-generated score.}
    \end{enumerate}

    We found that while reasoning increases alignment, it increases token costs far too greatly for it to be economically used in a d-RLAIF setting. So reasoning was excluded from being used in LLM-as-judge models.

\section{Training}
\label{appendix_training}

    Shown in Appendix Table \ref{tab:story-eval-wide-llama}, we conduct additional variations of post-training on Llama-3.1-8B-Instruct. Those trained by M-Prometheus achieve better performance when trained using R\textsubscript{O} and worse when trained using R\textsubscript{N} compared to those trained by Selene-1-mini. When the LLM-as-judge model is instructed to output a 5-point Likert score instead of the original 3-point and the group relative advantage is calculated from R\textsubscript{O5} instead of R\textsubscript{O}, the policy model's outputs improve at the Logical and Rational criteria but significantly degrade at the Complete\textsubscript{N} criteria. We also additionally run GRPO with ROUGE-L to train Llama 3.1-8B-instruct and find it worsens the performance of the model.

    \setlength{\tabcolsep}{3pt}
    \begin{table*}
      \centering
      \footnotesize
        \begin{tabular}{lllllllll}
          \toprule
          \multicolumn{1}{c}{\textbf{Author}} &
          \multicolumn{1}{c}{\textbf{Logical}} &
          \multicolumn{1}{c}{\textbf{Rational}} &
          \multicolumn{1}{c}{\makecell{\textbf{Complete}\textsubscript{N}}} &
          \multicolumn{1}{c}{\makecell{\textbf{min}\textsubscript{LRC}}} &
          \multicolumn{1}{c}{\textbf{Narrativity}} &
          \multicolumn{1}{c}{\textbf{BLEU-4}} &
          \multicolumn{1}{c}{\textbf{ROUGE-L}} \\
          \midrule
          \multicolumn{8}{@{}l}{\textbf{\textsc{Llama-3.1-8B-Instruct}}} \\
          BASE                         & 2.735 & 2.385 & 2.528 & 2.080 & 4.347 & 0.340 & 0.502 \\
          few-shot                     & 2.768 & 2.615 & 2.505 & 2.213 & 4.303 & 0.306 & 0.473 \\
          SFT                          & 2.320 & 2.501 & 2.934 & 2.260 & 4.692 & 0.810 & 0.835 \\
          GRPO\textsubscript{ROUGE-L}                          & 2.007 & 2.212 & \textbf{2.938} & 1.952 & 4.628 & \textbf{0.855} & \textbf{0.863} \\
          RRR\textsubscript{O}-Prometheus  & 2.801 & 2.825 & 2.594 & 2.400 & 4.534 & 0.179 & 0.386 \\
          RRR\textsubscript{O5}-Prometheus & \textbf{2.917} & \textbf{2.943} & 2.059 & 2.019 & 4.53 & 0.007 & 0.194 \\
          RRR\textsubscript{N}-Prometheus  & 2.655 & 2.717 & 2.246 & 2.058 & 4.219 & 0.163 & 0.357 \\
          RRR\textsubscript{O}-Selene      & 1.820 & 2.080 & 2.779 & 1.761 & 4.489 & 0.174 & 0.458 \\
          RRR\textsubscript{N}-Selene      & 2.785 & 2.860 & 2.740 & \textbf{2.554} & \textbf{4.733} & 0.087 & 0.299 \\
          \bottomrule
        \end{tabular}
      \caption{Evaluation of post-trained LLMs using the test split (n=1871) from TimeTravel dataset using Gemini-3-Flash. Additional models trained on Llama-3.1.}
      \label{tab:story-eval-wide-llama}
    \end{table*}

    \begin{figure*}[t!]
        \centering
        
        \begin{minipage}[t]{0.48\textwidth}
            \centering
            \includegraphics[width=0.9\linewidth]{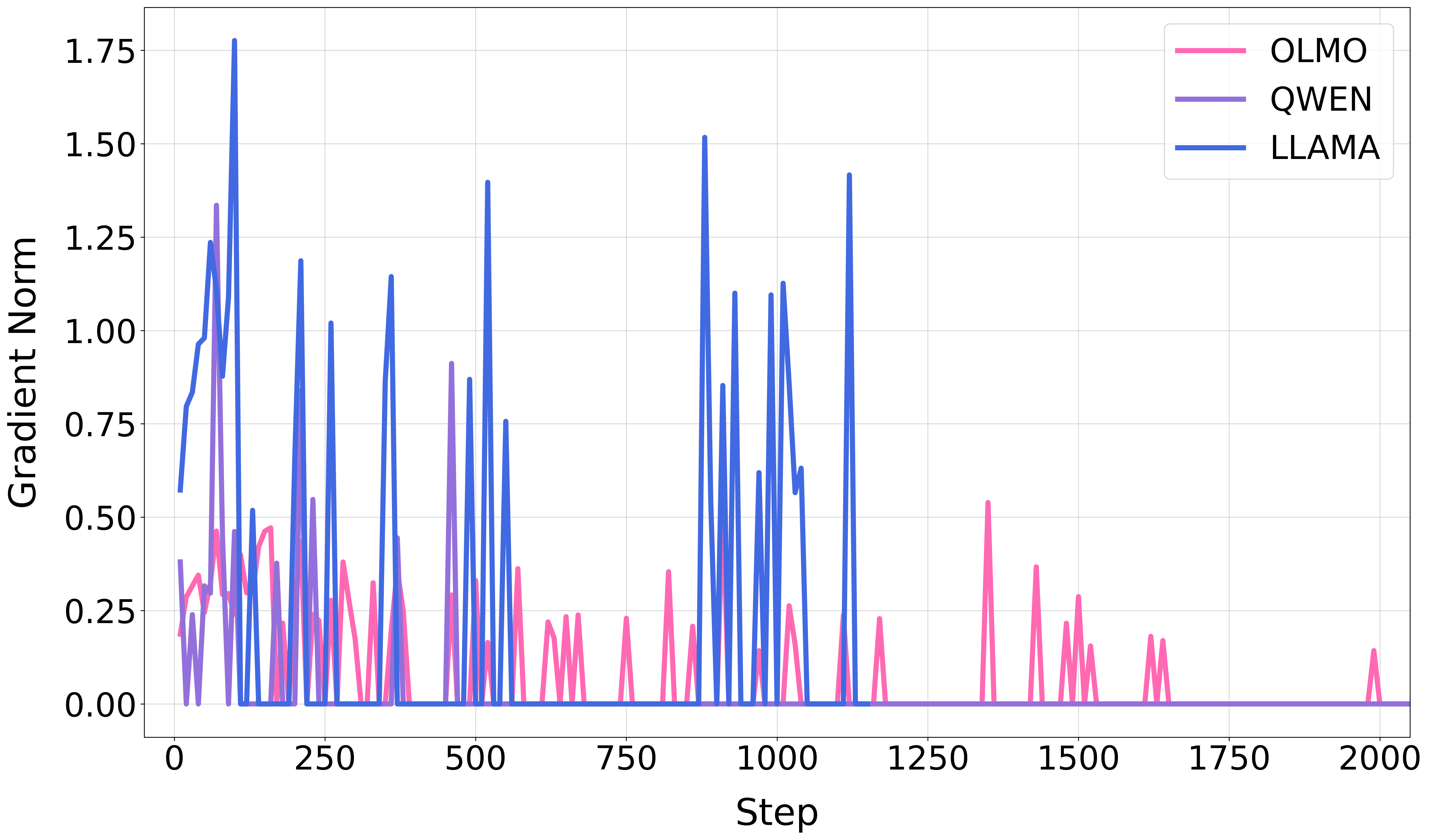}
            \caption{Policy model gradient norm during d-RLAIF using the reward signal R\textsubscript{O} generated by Selene-1-mini.}
            \label{fig:ro_selene_gradient}
        \end{minipage}\hfill
        \begin{minipage}[t]{0.48\textwidth}
            \centering
            \includegraphics[width=0.9\linewidth]{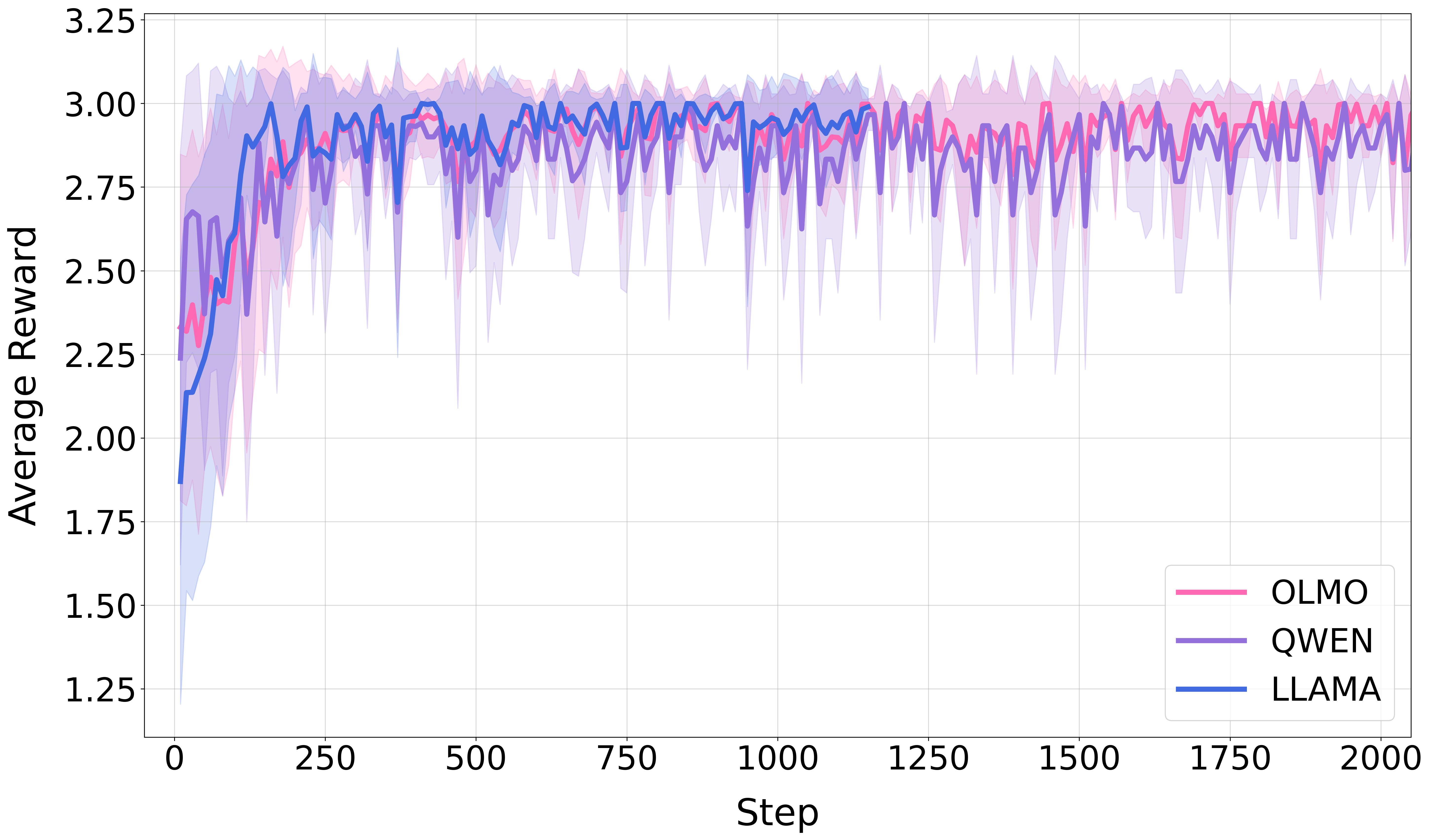}
            \caption{Mean and standard deviation of the reward signal R\textsubscript{O} generated by Selene-1-mini during d-RLAIF.}
            \label{fig:ro_selene}
        \end{minipage}
        
        \bigskip 
        
        \begin{minipage}[t]{0.48\textwidth}
            \centering
            \includegraphics[width=0.9\linewidth]{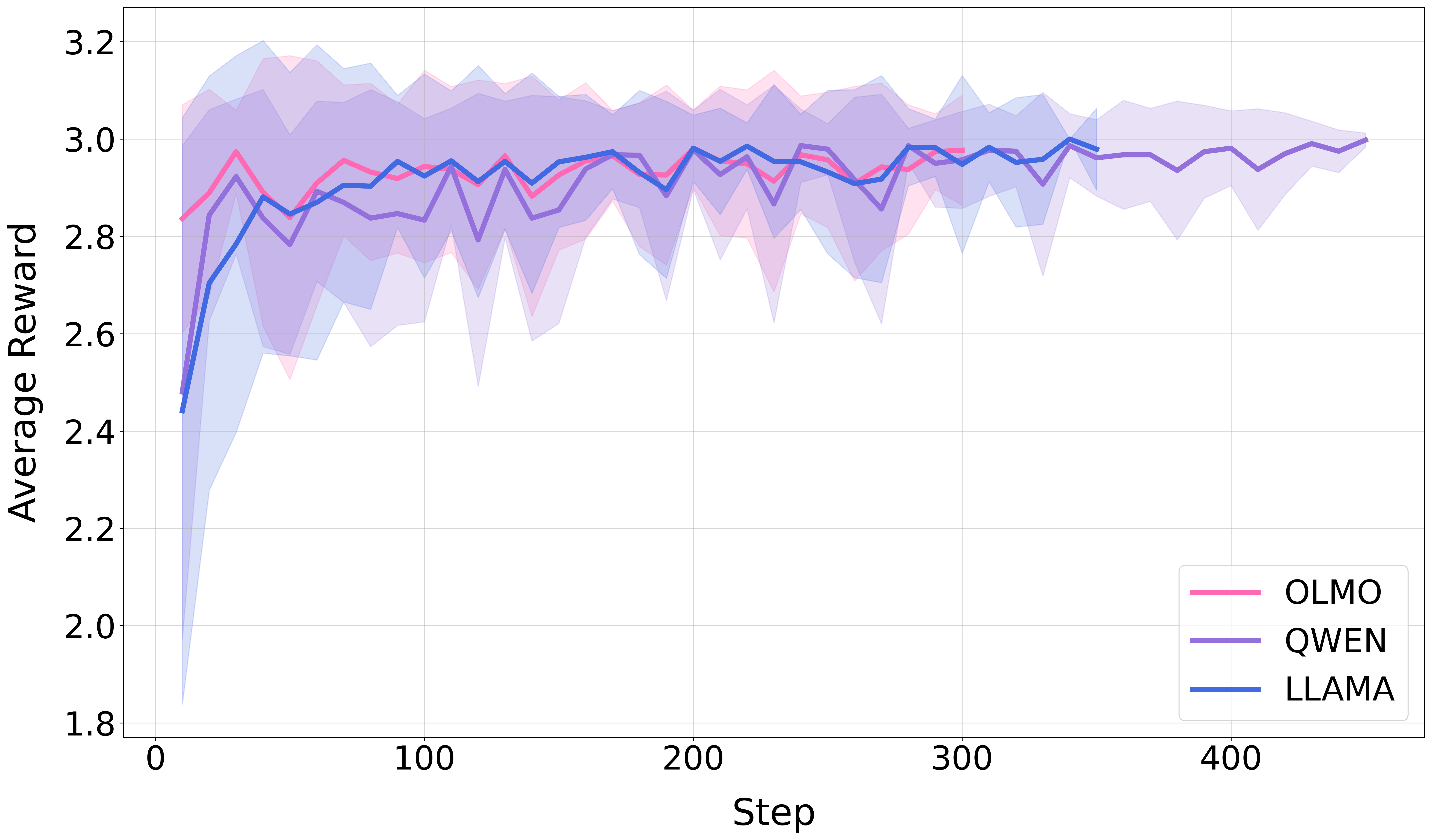}
            \caption{Mean and standard deviation of the reward signal R\textsubscript{O} generated by M-Prometheus during d-RLAIF.}
            \label{fig:ro_prometheus}
        \end{minipage}\hfill
        \begin{minipage}[t]{0.48\textwidth}
            \centering
            \includegraphics[width=0.9\linewidth]{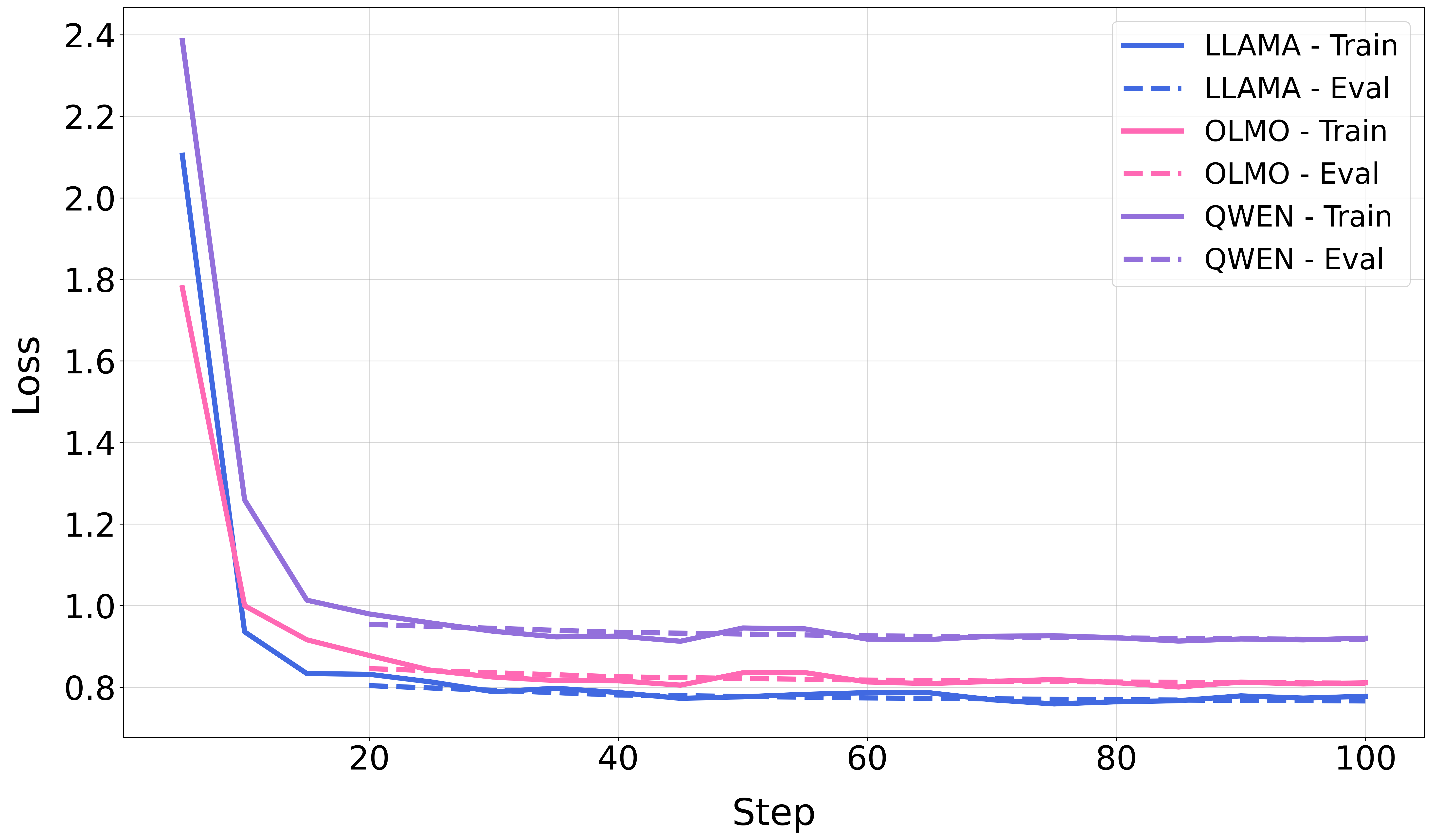}
            \caption{Training and evaluation loss during SFT.}
            \label{fig:sft_loss}
        \end{minipage}
        
    \end{figure*}
    
\clearpage
\onecolumn
\section{Significance Testing}
\label{sec:significance_testing}

\begin{table}[h]
  \centering
  \setlength{\tabcolsep}{1pt}
  \begin{tabular}{ll lllll}
    \toprule
    \textbf{Model} & 
    \textbf{Method} & 
    \textbf{Logical} &
    \textbf{Rational} &
    \makecell{\textbf{Complete}\textsubscript{N}} &
    \makecell{\textbf{min}\textsubscript{LRC}} &
    \textbf{Narrativity} \\
    \midrule
    
    \multirow{5}{*}{\textbf{\textsc{Llama-3.1-8B-Inst.}}} 
    & BASE                 & 2.735* & 2.385* & 2.528* & 2.080* & 4.347* \\
    & few-shot             & 2.768  & 2.615* & 2.505* & 2.213* & 4.303* \\
    & SFT                  & 2.320* & 2.501* & \textbf{2.934}* & 2.260* & 4.692  \\
    & RRR\textsubscript{O}-Prometheus & \textbf{2.801} & 2.825* & 2.594* & 2.400* & 4.534* \\
    & RRR\textsubscript{N}-Selene     & 2.785 \textsuperscript{Base} & \textbf{2.860} \textsuperscript{Base} & 2.740 \textsuperscript{Base} & \textbf{2.554} \textsuperscript{Base} & \textbf{4.733} \textsuperscript{Base} \\
    \bottomrule
  \end{tabular}
  
  \caption{Evaluation of Llama-3.1-8B variants using the test split ($n=1871$) from the TimeTravel dataset. Values marked with * indicate a statistically significant difference ($p < 0.05$ via Wilcoxon signed-rank test) compared to the \textbf{RRR\textsubscript{N}-Selene} baseline. Bolded values represent the best performance for each metric.}
  \label{tab:llama-eval-significance}
\end{table}

\begin{table}[h]
  \centering
  \begin{tabular}{ll lllll}
    \toprule
    \textbf{Model} & 
    \textbf{Method} & 
    \textbf{Logical} &
    \textbf{Rational} &
    \makecell{\textbf{Complete}\textsubscript{N}} &
    \makecell{\textbf{min}\textsubscript{LRC}} &
    \textbf{Narrativity} \\
    \midrule
    
    \multirow{5}{*}{\textbf{\textsc{Qwen-3-8B}}} 
    & BASE                 & 2.429* & 2.132* & 2.759* & 1.940* & 4.551* \\
    & few-shot             & 2.543* & 2.219* & 2.686  & 2.006* & 4.559* \\
    & SFT                  & 2.402* & 2.532* & \textbf{2.925}* & 2.317* & 4.693* \\
    & RRR\textsubscript{O}-Prometheus & 2.717* & 2.336* & 2.566* & 2.024* & 4.356* \\
    & RRR\textsubscript{N}-Selene     & \textbf{2.781} \textsuperscript{Base} & \textbf{2.890} \textsuperscript{Base} & 2.705 \textsuperscript{Base} & \textbf{2.516} \textsuperscript{Base} & \textbf{4.894} \textsuperscript{Base} \\
    \bottomrule
  \end{tabular}
  
  \caption{Evaluation of Qwen-3-8B variants using the test split ($n=1871$) from the TimeTravel dataset. Values marked with * indicate a statistically significant difference ($p < 0.05$ via Wilcoxon signed-rank test) compared to the \textbf{RRR\textsubscript{N}-Selene} baseline. Bolded values represent the best performance for each metric within the model family.}
  \label{tab:qwen-eval-significance}
\end{table}

\begin{table}[h]
  \centering
  \begin{tabular}{ll lllll}
    \toprule
    \textbf{Model} & 
    \textbf{Method} & 
    \textbf{Logical} &
    \textbf{Rational} &
    \makecell{\textbf{Complete}\textsubscript{N}} &
    \makecell{\textbf{min}\textsubscript{LRC}} &
    \textbf{Narrativity} \\
    \midrule
    
    \multirow{5}{*}{\textbf{\textsc{Olmo-3-7B-Inst.}}} 
    & BASE                 & 2.762  & 2.812* & 2.400* & 2.222* & 4.328* \\
    & few-shot             & 2.745* & 2.837* & 2.377* & 2.208* & 4.415* \\
    & SFT                  & 2.295* & 2.441* & \textbf{2.910}* & 2.211* & 4.647* \\
    & RRR\textsubscript{O}-Prometheus & 2.770  & 2.857* & 2.491* & 2.325* & 4.460* \\
    & RRR\textsubscript{N}-Selene     & \textbf{2.793} \textsuperscript{Base} & \textbf{2.893} \textsuperscript{Base} & 2.595 \textsuperscript{Base} & \textbf{2.434} \textsuperscript{Base} & \textbf{4.899} \textsuperscript{Base} \\
    \bottomrule
  \end{tabular}
  
  \caption{Evaluation of Olmo-3-7B variants using the test split ($n=1871$) from the TimeTravel dataset. Values marked with * indicate a statistically significant difference ($p < 0.05$ via Wilcoxon signed-rank test) compared to the \textbf{RRR\textsubscript{N}-Selene} baseline. Bolded values represent the best performance for each metric within the model family.}
  \label{tab:olmo-eval-significance}
\end{table}

\section{Example Annotations (with Additional Reasoning/Rationale)}
\label{sec:inter_annotator_diff}

\begin{table}[H]
  \renewcommand{\arraystretch}{0.9}
  \footnotesize
  \centering
  \setlength{\tabcolsep}{3pt}
  \begin{tabularx}{\textwidth}{@{}X@{}}
  \toprule

  \textbf{Narrativity} (\hleq{Equilibrium}→\hldis{Disruption}→\hlrec{Recognition}→\hlatt{Attempt}→\hlnew{New Equilibrium}) \\[2pt]

  \textbf{A:} (4)
  \hleq{My nephew had a bad back one day.}
  He took some ibuprofen and was fine. The pills were never even opened, and his back felt better by the next day.
  \\[4pt]

  \textbf{B:} (4)
  \hldis{My nephew had a bad back one day.}
  \hlatt{He took some ibuprofen and was fine.}
  \hlnew{The pills were never even opened, and his back felt better by the next day.}
  \\[4pt]

  \textbf{C:} (4)
  \hldis{My nephew had a bad back one day.}
  \hlatt{He took some ibuprofen}
  \hlnew{was fine.}
  The pills were never even opened, and
  \hlnew{his back felt better by the next day.}
  \\[6pt]

  \toprule

  \textbf{Logical} \\[2pt]
  \textbf{A:} (1) It's illogical and inconceivable that the nephew took ibuprofen yet the pills were never opened. \\
  \textbf{B:} (3) The edited text is logically coherent. Taking ibuprofen and leaving the prescription pills unopened is fully consistent. \\
  \textbf{C:} (3) The story does not present inconceivable scenarios. Whilst at first there appears to be a logical inconsistency, “the pills” do not necessarily refer to ibuprofen. \\

  \midrule

  \textbf{Rational} \\[2pt]
  \textbf{A:} (1) There's causal and contextual disconnection between taking ibuprofen and the pills being never opened. \\
  \textbf{B:} (3) The edited text is rationally interpretable. The progression from pain to treatment to recovery is clear. \\
  \textbf{C:} (1) The story cannot rationally be interpreted using Todorov's stages in its entirety because there is a causal disconnection; the most likely interpretation is that “the pills” refer to the ibuprofen in sentence two. \\

  \midrule

  \textbf{Complete\textsubscript{N}} \\[2pt]
  \textbf{A:} (1) There are no recognisable disruptions. The edited text is not as much of a narrative as the original. \\
  \textbf{B:} (3) The edited text is complete. It preserves the same broad Todorovian stage structure as the original. \\
  \textbf{C:} (3) All stages of Todorov's theory presented in the original text are present in the rewritten story. \\

  \bottomrule
  \end{tabularx}
  \caption{An example of valid annotator disagreement when tagging and evaluating stories.}
  \label{tab:inter_annotator_back_pain}
\end{table}

\begin{table}[h]
  \renewcommand{\arraystretch}{0.9}
  \footnotesize
  \centering
  \setlength{\tabcolsep}{3pt}
  \begin{tabularx}{\textwidth}{@{}X@{}}
  \toprule
  
  \textbf{Narrativity} (\hleq{Equilibrium}→\hldis{Disruption}→\hlrec{Recognition}→\hlatt{Attempt}→\hlnew{New Equilibrium}) \\[2pt]
  
  \textbf{A:} (5)
  \hldis{I started using coupons.}
  \hlrec{I realized it was too much work, with no real savings.}
  \hlatt{I still collected coupons from the Sunday paper, emails and snail mail, but I didn't put much effort into researching stores that double the value of coupons.}
  \hlnew{In the end, I didn't clip, tear, or cut any coupons, and I didn't end up saving any money last week.}
  \\[4pt]
  
  \textbf{B:} (5)
  \hleq{I wanted to see if i could save some money so I started using coupons.}
  \hlrec{I realized it was too much work, }
  \hldis{with no real savings.}
  \hlatt{I still collected coupons from the Sunday paper, emails and snail mail, but I didn't put much effort into researching stores that double the value of coupons.}
  \hlnew{In the end, I didn't clip, tear, or cut any coupons, and I didn't end up saving any money last week.}
  \\[4pt]
  
  \textbf{C:} (4)
  \hldis{I wanted to see if i could save some money so I started using coupons.}
  \hlrec{I realized it was too much work, with no real savings.}
  I still collected coupons from the Sunday paper, emails and snail mail, but I didn't put much effort into researching stores that double the value of coupons.
  \hlnew{In the end, I didn't clip, tear, or cut any coupons, and I didn't end up saving any money last week.}
  \\[6pt]
  
  \toprule
  
  \textbf{Logical} \\[2pt]
  \textbf{A:} (1) It is illogical that the narrator would keep collecting coupons after realising that it was too much work with no real savings. \\
  \textbf{B:} (3) The edited text is logically coherent. The speaker tries coupons, recognises the effort outweighs the savings, reduces that effort, and ends without saving money. \\
  \textbf{C:} (3) The events of the story progress without inconceivable scenarios. A logical inconsistency could be interpreted between sentences three and four... However, this inconsistency could be explained if we viewed the time of narration as a flashback. \\
  
  \midrule
  
  \textbf{Rational} \\[2pt]
  \textbf{A:} (1) There is no rational causal connection between realising that coupons are too much work and continuing to collect coupons. \\
  \textbf{B:} (3) The edited text is rationally interpretable. The causal progression from initial goal to failed outcome is clear. \\
  \textbf{C:} (3) The events progress rationally: the character wants to see if they can save money with coupons, they realise it is too much work, then stop attempting to do so. \\
  
  \midrule
  
  \textbf{Complete\textsubscript{N}} \\[2pt]
  \textbf{A:} (3) It has a narrative structure of disruption, recognition, attempt and equilibrium. Even though it doesn't make sense, it's as much of a narrative as the original. \\
  \textbf{B:} (3) The edited text is complete. It includes disruption, recognition, attempt, and a new equilibrium, and is at least as stage-complete as the original. \\
  \textbf{C:} (1) The rewritten story is less complete than the original because it omits the crucial “attempt” stage. Even though the rewritten story includes as many Todorovian stages, it is not as narratively complete because character agency isn't represented as fully. \\
  
  \bottomrule
  \end{tabularx}
  \caption{An example of valid annotator disagreement when tagging and evaluating stories.}
  \label{tab:inter_annotator_coupons}
\end{table}
\begin{table}[h]
    \renewcommand{\arraystretch}{0.9}
    \footnotesize
    \centering
    \setlength{\tabcolsep}{3pt}
    \begin{tabularx}{\textwidth}{@{}X@{}} 
    \toprule
    \textbf{Narrativity} (\hleq{Equilibrium}→\hldis{Disruption}→\hlrec{Recognition}→\hlatt{Attempt}→\hlnew{New Equilibrium})\\ 
    \textbf{A:}
    (4)
    \hleq{Evan was ready to buy himself a bicycle.} 
    \hldis{He went to the store but forgot his money.} 
    \hlatt{Then he looked around some more, trying to find something else he could afford or a way to get home.}
    \\
    \textbf{B:} 
    (4)
    \hldis{Evan was ready to buy himself a bicycle.} 
    \hlatt{He went to the store} 
    \hldis{but forgot his money.} 
    \hlnew{Then he looked around some more, trying to find something else he could afford or a way to get home.}
    \\
    \textbf{C:} 
    (3)
    \hldis{Evan was ready to buy himself a bicycle.} 
    \hlatt{He went to the store but forgot his money.} 
    Then he looked around some more, trying to find something else he could afford or a way to get home.
    \\
    \toprule 
    
    \textbf{Logical} \\ 
    \textbf{A:} (3) I interpret ‘but forgot his money’ as Evan forgetting a large sum of money intended for buying a bicycle. Evan may still have the usual money he carries every day, and is able to logically find something else he could afford. \\
    \textbf{B:} (3) The edited text is logical. Forgetting his money and then looking for an alternative or a way home is internally consistent. \\
    \textbf{C:} (1) There is a logical impossibility in the story because: 1. Evan forgets his money, but 2. proceeds to look around the store for something he can “afford,” which would be nothing. \\
    \midrule
    
    \textbf{Rational} \\ 
    \textbf{A:} (3) Looking for an alternative after forgetting the money to buy the bicycle is causally and contextually connected. \\
    \textbf{B:} (1) The edited text is irrational in Todorovian terms. The attempt is present, but the narrative does not clearly resolve the disruption or establish a justified new equilibrium. \\
    \textbf{C:} (1) The events of the story are causally disconnected. \\
    \midrule
    
    \textbf{Complete\textsubscript{N}} \\ 
    \textbf{A:} (1) The edited text does not reach a new equilibrium whereas the original text does: “Then he rode home!” so it is not as complete of a narrative as the original. \\
    \textbf{B:} (3) The edited text is complete. It includes disruption, attempt, and a form of new equilibrium, which matches the broad stage structure of the original. \\
    \textbf{C:} (1) The original text presented a disruption in Evan's desire to buy a bicycle; an 'attempt at doing so in looking around the store; and a 'new equilibrium' after he purchases the bicycle he initially desired. \\
    \bottomrule
    \end{tabularx}
    \caption{An example of valid annotator disagreement when tagging and evaluating stories.}
    \label{tab:inter_annotator_bicycle2}
\end{table}

\begin{table}[h]
  \renewcommand{\arraystretch}{0.9}
  \footnotesize
  \centering
  \setlength{\tabcolsep}{3pt}
  \begin{tabularx}{\textwidth}{@{}X@{}}
  \toprule
  
  \textbf{Narrativity} (\hleq{Equilibrium}→\hldis{Disruption}→\hlrec{Recognition}→\hlatt{Attempt}→\hlnew{New Equilibrium})\\[2pt]
  
  \textbf{A:} (1)
  \hleq{I had to go to Boston for a work trip.}
  \hleq{It was a boring trip filled with conferences and no fun.}
  \hleq{We attended several conferences and had no memorable experiences,}
  \hleq{but I learned a lot for my career that I will carry with me.}
  \\[4pt]
  
  \textbf{B:} (4)
  \hldis{I had to go to Boston for a work trip.}
  \hlrec{It was a boring trip filled with conferences and no fun.}
  We attended several conferences and had no memorable experiences,
  \hlnew{but I learned a lot for my career that I will carry with me.}
  \\[4pt]
  
  \textbf{C:} (4)
  \hldis{I had to go to Boston for a work trip}
  \hlrec{It was a boring trip filled with conferences and no fun.}
  \hlrec{We attended several conferences and had no memorable experiences, }
  but
  \hlnew{I learned a lot for my career that I will carry with me}
  \\[6pt]
  
  \toprule
  
  \textbf{Logical} \\[2pt]
  \textbf{A:} (3) The scenario is conceivable and breaks no logical rules. “boring/unmemorable” and “learn a lot” are not mutually exclusive. \\
  \textbf{B:} (1) The edited text contains a weak internal contradiction: the trip is described as having “no memorable experiences,” yet it also results in a significant and lasting professional benefit. Because this shift is not sufficiently explained, the narrative’s internal logic is weakened. \\
  \textbf{C:} (3) At first, it seems there is a logical impossibility because the narrator, after having an uneventful trip “filled with [...] no fun” and “no memorable experiences,” expresses in the last line that they learnt a lot. However, the conjunction “but” suggests that even though they had a bad time, the experience nevertheless taught them something. Bad or uneventful experiences can also be informative. \\
  
  \midrule
  
  \textbf{Rational} \\[2pt]
  \textbf{A:} (3) Looking for an alternative after forgetting the money to buy the bicycle is causally and contextually connected. \\ 
  \textbf{B:} (1) Although the text signals disruption and ends in a new equilibrium, it does not rationally develop the causal steps between them. The transition from a boring, uneventful trip to a meaningful career lesson lacks narrative grounding. \\
  \textbf{C:} (3) Again, there appears to be a causal disconnection within the last sentence (from recognition to equilibrium). However, the word “but” makes the story rational. If the conjunction were “and”, the story would be causally disconnected. \\
  
  \midrule
  
  \textbf{Complete\textsubscript{N}} \\[2pt]
  \textbf{A:} (1) There is no disruption in the edited text. It is not as complete as the original narrative, which has disruption and new equilibrium. \\
  \textbf{B:} (1) The edited text is not fully complete because it excludes the Attempt stage. Without that stage, the sequence of Todorovian development is reduced, making the edited version less narratively complete than the original. \\
  \textbf{C:} (3) Even though the fourth stage of Todorov’s theory seems more important to create a complete narrative, the rewritten story feels just as complete because of the final clause “but I learned a lot for my career that I will carry with me.” \\
  
  \bottomrule
  \end{tabularx}
  \caption{An example of valid annotator disagreement when tagging and evaluating stories.}
  \label{tab:inter_annotator_boston_full}
\end{table}

\begin{table}[h]
  \renewcommand{\arraystretch}{0.9}
  \footnotesize
  \centering
  \setlength{\tabcolsep}{3pt}
  \begin{tabularx}{\textwidth}{@{}X@{}}
  \toprule

  \textbf{Narrativity} (\hleq{Equilibrium}→\hldis{Disruption}→\hlrec{Recognition}→\hlatt{Attempt}→\hlnew{New Equilibrium}) \\[2pt]

  \textbf{A:} (5)
  \hleq{In the last mile of a marathon only two runners were in contention.}
  \hldis{One runner got cramps on the ground and tripped the other.}
  \hlatt{They both rested on the ground.}
  They rested for a few second \textit{[sic]}
  \hlnew{before they drop out.}
  \\[4pt]

  \textbf{B:} (5)
  \hleq{In the last mile of a marathon only two runners were in contention.}
  \hldis{They both got cramps and had to drop out of the race.}
  \hldis{One runner got cramps on the ground and tripped the other.}
  \hlrec{They both rested on the ground.}
  \hlatt{They rested for a few second \textit{[sic]}}
  \hlnew{before they drop out.}
  \\[4pt]

  \textbf{C:} (3)
  \hleq{In the last mile of a marathon only two runners were in contention.}
  \hldis{They both got cramps and had to drop out of the race.}
  One runner got cramps on the ground and tripped the other. They both rested on the ground. They rested for a few second \textit{[sic]} before they drop out.
  \\[6pt]

  \toprule

  \textbf{Logical} \\[2pt]
  \textbf{A:} (3) Conceivable. The events are not narrated in chronological order. “They both got cramps and had to drop out” is a summary of the explanations that follow. \\
  \textbf{B:} (1) The edited text is not logical. The claim that one runner “got cramps on the ground and tripped the other” is internally unclear and weakens the event sequence. \\
  \textbf{C:} (3) The events of the story progress logically, free from internal contradictions; the two runners both get cramps, drop to the floor, and will eventually drop out. \\

  \midrule

  \textbf{Rational} \\[2pt]
  \textbf{A:} (3) Everything is contextually and causally connected even if it's not in chronological order. \\
  \textbf{B:} (1) The edited text is not rational. The causal progression from cramping to tripping to resting to dropping out is not presented in a fully coherent way. \\
  \textbf{C:} (3) All events can be rationally interpreted using the stages. The last two sentences are an elaboration of the disruption that I tagged in sentence two. \\

  \midrule

  \textbf{Complete\textsubscript{N}} \\[2pt]
  \textbf{A:} (3) It is just as complete of a narrative as the original. Cramps are the disruption, resting is the attempt to resolve, and dropping out is the new equilibrium. \\
  \textbf{B:} (3) The edited text is complete in stage coverage. It still presents equilibrium, disruption, recognition, attempt, and a final outcome. \\
  \textbf{C:} (1) The rewritten story omits an attempt at new equilibrium because there is no activity to overcome the obstacle of falling to the ground with cramps. \\

  \bottomrule
  \end{tabularx}
  \caption{An example of valid annotator disagreement when tagging and evaluating stories.}
  \label{tab:inter_annotator_marathon}
\end{table}

\clearpage
\section{Example Outputs}
\label{sec:example_outputs}


\begin{table}[h]
\centering
\footnotesize
\setlength{\tabcolsep}{4pt}
\renewcommand{\arraystretch}{1}

\begin{tabularx}{\textwidth}{@{}p{7em}X p{0.5em} p{0.5em} p{0.5em} p{0.5em} p{0.5em}@{}}
\toprule
\textbf{Source} & \textbf{Text} & \textbf{L} & \textbf{R} & \textbf{C} & \textbf{M} & \textbf{N} \\
\midrule

\rowcolor{black!5}\multicolumn{7}{@{}l}{\textbf{Original Story}}\\
\midrule
Premise & Max and May were at a tavern. & & & & & \\
Initial & They saw the dartboard and challenged each other to a game. & & & & & \\
Original Ending & Max played very well. But May thru even better than him. May won the game. & & & & & \\

\midrule
\rowcolor{black!5}\multicolumn{7}{@{}l}{\textbf{Counterfactual Retelling}}\\
\midrule
Premise & Max and May were at a tavern. & & & & & \\
Counterfactual & They didn't play darts there because they were not competitive. & & & & & \\
Edited Ending & ??? & & & & & \\

\midrule

\multicolumn{7}{@{}l}{\textbf{Human Ground Truth}}\\
Crowd Worker & Max played pool very well. But May thought she was even better than him. May won the game, but neither really cared who won. & 3 & 3 & 3 & 3 & 5 \\

\midrule
\multicolumn{7}{@{}l}{\textbf{Olmo}}\\
Base & Max and May decided not to play since they weren't competitive. & 3 & 3 & 1 & 1 & 1 \\
Few-Shot & Max and May talked and laughed instead of playing. They enjoyed the atmosphere together. & 3 & 3 & 3 & 3 & 1 \\
SFT & Max played very well. But May threw even better than him. May won the game. & 1 & 1 & 3 & 1 & 1 \\
RRR\textsubscript{O} & They left the tavern satisfied with their decision not to compete. & 3 & 3 & 3 & 3 & 1 \\
RRR\textsubscript{N} & They decided to sing a song instead. The tavern erupted in laughter and applause—both agreed it was far more fun than competition. & 3 & 3 & 3 & 3 & 5 \\

\midrule
\multicolumn{7}{@{}l}{\textbf{Llama}}\\
Base & Max and May were at a tavern. They saw the dartboard but decided not to play because they were not competitive. & 3 & 1 & 1 & 1 & 1 \\
Few-Shot & They decided to skip the dartboard and instead ordered a couple of beers to relax and enjoy the evening. & 3 & 3 & 1 & 1 & 1 \\
SFT & Max played very well at cards. But May threw even better than him. May won the game. & 1 & 3 & 3 & 1 & 5 \\
RRR\textsubscript{O} & They ended up chatting over a pint instead, enjoying each other's company without the competitive edge that darts had brought out in them. & 3 & 3 & 1 & 1 & 5 \\
RRR\textsubscript{N} & They ordered a round and struck up a conversation. Enjoying each other's company, they decided to share a story. Max won over May with his tale. & 3 & 3 & 3 & 3 & 5 \\

\midrule
\multicolumn{7}{@{}l}{\textbf{Qwen}}\\
Base & Max played very well. But May threw even better than him. May won the game. & 1 & 1 & 3 & 1 & 1 \\
Few-Shot & They didn't play darts there because they were not competitive. Instead, they talked about their favorite books and shared stories about their travels. & 3 & 1 & 1 & 1 & 1 \\
SFT & Max played very well at darts. But May threw even better than him. May won the game. & 1 & 1 & 3 & 1 & 5 \\
RRR\textsubscript{O} & They decided to play a different game instead, one that allowed them to enjoy each other's company without competition. & 3 & 3 & 1 & 1 & 4 \\
RRR\textsubscript{N} & They ordered drinks, then debated chess. Max won, but May vowed to train. They vowed to rematch. & 3 & 3 & 3 & 3 & 5 \\

\midrule
\multicolumn{7}{@{}l}{\textbf{Codex}}\\
Code-Prompt & Max played very well. But May thru even better than him. They had a good time. & 1 & 1 & 1 & 1 & 1 \\

\bottomrule
\end{tabularx}

\caption{Example outputs from the test set. Ratings are given by LLM-as-judge (Gemini-3-Flash): L=Logical, R=Rational, C=Complete\textsubscript{N}, M=min\textsubscript{LRC}, N=Narrativity.}
\label{tab:consolidated-merged}
\end{table}


\begin{table}[h]
\centering
\footnotesize
\setlength{\tabcolsep}{4pt}
\renewcommand{\arraystretch}{1}

\begin{tabularx}{\textwidth}{@{}p{7em}X p{0.5em} p{0.5em} p{0.5em} p{0.5em} p{0.5em}@{}}
\toprule
\textbf{Source} & \textbf{Text} & \textbf{L} & \textbf{R} & \textbf{C} & \textbf{M} & \textbf{N} \\
\midrule

\rowcolor{black!5}\multicolumn{7}{@{}l}{\textbf{Original Story}}\\
\midrule
Premise & Kimmy was on a trip to Paris. & & & & & \\
Initial & Her plane had just landed and she and her classmates her \textit{[sic]} excited. & & & & & \\
Original Ending & They couldn't wait to start seeing museums and churches! They spent lots of time shopping and seeing historical sites. Kimmy was very excited. & & & & & \\

\midrule
\rowcolor{black!5}\multicolumn{7}{@{}l}{\textbf{Counterfactual Retelling}}\\
\midrule
Premise & Kimmy was on a trip to Paris. & & & & & \\
Counterfactual & Kimmy's plane crashed over the ocean and killed everyone on board. & & & & & \\
Edited Ending & ??? & & & & & \\

\midrule

\multicolumn{7}{@{}l}{\textbf{Human Ground Truth}}\\
Crowd Worker & Kimmy had just wanted to see the museums and churches! Her parents spent a lot of time planning her funeral. Kimmy was dead. & 3 & 3 & 3 & 3 & 5 \\

\midrule
\multicolumn{7}{@{}l}{\textbf{Olmo}}\\
Base & Given the tragic counterfactual, there is no living person to experience excitement or see museums and churches, so the premise cannot be fulfilled in the same way as the original ending. & 3 & 3 & 1 & 1 & 3 \\
Few-Shot & There is no edited ending, as the counterfactual scenario is a tragic accident with no possible positive or alternative ending. & 3 & 3 & 1 & 1 & 3 \\
SFT & They couldn't wait to start seeing museums and churches! They spent lots of time shopping and seeing historical sites. Kimmy was very excited. & 1 & 1 & 3 & 1 & 1 \\
RRR\textsubscript{O} & Due to the tragic accident, the trip never happened, and Kimmy and her classmates never experienced Paris or its attractions. There were no museums, churches, shopping, or historical sites to see. Kimmy could never have been excited in the same way. & 3 & 3 & 3 & 3 & 4 \\
RRR\textsubscript{N} & Grief-stricken, the world mourned. No legacy remained—only memories in the waves. The tragedy ended hope; Paris dreams faded. The story closed with silent sorrow. & 3 & 3 & 3 & 3 & 5 \\

\midrule
\multicolumn{7}{@{}l}{\textbf{Llama}}\\
Base & They never made it to Paris. Kimmy's plane crashed over the ocean and there were no survivors. & 3 & 1 & 3 & 1 & 3 \\
Few-Shot & Kimmy's plane crashed over the ocean and killed her. & 3 & 3 & 1 & 1 & 3 \\
SFT & They couldn't wait to start seeing museums and churches! They spent lots of time shopping and seeing historical sites. Kimmy was very excited. & 1 & 1 & 3 & 1 & 1 \\
RRR\textsubscript{O} & The plane wreckage was found at the bottom of the ocean, a tragic reminder of the devastating crash that took Kimmy and her classmates' lives. & 3 & 3 & 3 & 3 & 4 \\
RRR\textsubscript{N} & Panicked, Kimmy clung to debris, the only survivor. She was rescued and returned home, re-evaluating life. Kimmy was forever changed. & 1 & 3 & 3 & 1 & 5 \\

\midrule
\multicolumn{7}{@{}l}{\textbf{Qwen}}\\
Base & There were no survivors, and Kimmy's adventure in Paris was cut short tragically. The classmates and everyone on board lost their lives in the crash. & 3 & 3 & 3 & 3 & 3 \\
Few-Shot & Kimmy's plane crashed over the ocean and killed everyone on board. There was no one left to see museums and churches, and no one to share the excitement with. Kimmy's trip ended tragically before it could begin. & 3 & 1 & 3 & 1 & 5 \\
SFT & They couldn't wait to start seeing museums and churches. They spent lots of time shopping and seeing historical sites. Kimmy was very excited about the trip. & 1 & 1 & 3 & 1 & 3 \\
RRR\textsubscript{O} & Unfortunately, Kimmy and her classmates were never able to explore Paris. The plane crash claimed their lives before they could experience the city's wonders. & 3 & 3 & 3 & 3 & 4 \\
RRR\textsubscript{N} & Miraculously, Kimmy survived and was rescued. She mourned her classmates but vowed to honor them. She traveled solo, explored Paris, and wrote a heartfelt tribute. Her story inspired many. & 1 & 3 & 3 & 1 & 5 \\

\midrule
\multicolumn{7}{@{}l}{\textbf{Codex}}\\
Code-Prompt & Kimmy was very sad. She was very sad. She was very sad. & 1 & 1 & 3 & 1 & 3 \\

\bottomrule
\end{tabularx}

\caption{Example outputs from the test set. Ratings are given by LLM-as-judge (Gemini-3-Flash): L=Logical, R=Rational, C=Complete\textsubscript{N}, M=min\textsubscript{LRC}, N=Narrativity.}
\label{tab:consolidated-merged}
\end{table}


\begin{table}[h]
\centering
\footnotesize
\setlength{\tabcolsep}{4pt}
\renewcommand{\arraystretch}{1}

\begin{tabularx}{\textwidth}{@{}p{7em}X p{0.5em} p{0.5em} p{0.5em} p{0.5em} p{0.5em}@{}}
\toprule
\textbf{Source} & \textbf{Text} & \textbf{L} & \textbf{R} & \textbf{C} & \textbf{M} & \textbf{N} \\
\midrule

\rowcolor{black!5}\multicolumn{7}{@{}l}{\textbf{Original Story}}\\
\midrule
Premise & Amy's class went on a field trip to see A Mid Summer Night's Dream. & & & & & \\
Initial & They arrived at the theater and found their seats. & & & & & \\
Original Ending & Amy was in awe of the actors' discipline and talent. After the play she was able to speak with the cast. Amy was thrilled. & & & & & \\

\midrule
\rowcolor{black!5}\multicolumn{7}{@{}l}{\textbf{Counterfactual Retelling}}\\
\midrule
Premise & Amy's class went on a field trip to see A Mid Summer Night's Dream. & & & & & \\
Counterfactual & They forgot their tickets in the classroom. & & & & & \\
Edited Ending & ??? & & & & & \\

\midrule

\multicolumn{7}{@{}l}{\textbf{Human Ground Truth}}\\
Crowd Worker & Amy was in awe at their stupidity. She saw the play the next week and then she was able to speak with the cast. Amy was thrilled. & 3 & 3 & 3 & 3 & 5 \\

\midrule
\multicolumn{7}{@{}l}{\textbf{Llama}}\\
Base & They had to go back to the classroom to get the tickets, but were able to find their seats just in time for the start of the play. Amy was in awe of the actors' discipline and talent. After the play, they were unable to speak with the cast because they had to rush back to the classroom to pick up their forgotten items, but Amy was still thrilled to have seen the performance. & 3 & 3 & 3 & 3 & 5 \\
Few-Shot & Amy was frustrated that they couldn't get in to see the play. They had to go back to the classroom to get the tickets. After finally getting in, Amy was in awe of the actors' discipline and talent. However, she was disappointed that they missed the beginning of the play. & 3 & 3 & 3 & 3 & 5 \\
SFT & Amy was in awe of the actors discipline and talent. After the play she was able to speak with the cast. Amy was thrilled. & 2 & 1 & 3 & 3 & 3 \\
RRR\textsubscript{O} & Panicked, they rushed back to the classroom to retrieve their tickets, but unfortunately, the class had already left for the theater. Disappointed, they decided to watch from outside the theater, still managing to catch a glimpse of the performance and hearing the sound of the actors' voices carrying through the windows. & 1 & 3 & 1 & 1 & 5 \\
RRR\textsubscript{N} & Panicked, they rushed back to school. Luckily, they found the tickets. They arrived just in time, and Amy was amazed by the performance. She met the cast backstage, beaming with joy. & 3 & 3 & 3 & 3 & 5 \\

\midrule
\multicolumn{7}{@{}l}{\textbf{Qwen}}\\
Base & Amy was in awe of the actors' discipline and talent. After the play, she realised she had forgotten her tickets in the classroom and couldn't speak with the cast. Amy was disappointed. & 1 & 1 & 3 & 1 & 5 \\
Few-Shot & They forgot their tickets in the classroom. Amy was disappointed and worried they would miss the play. After a frantic search, they found the tickets just in time to enter the theater. & 3 & 1 & 3 & 1 & 5 \\
SFT & Amy was in awe of the actors discipline and talent. After the play she was able to speak with the cast. Amy was thrilled. & 2 & 1 & 3 & 1 & 5 \\
RRR\textsubscript{O} & Amy was frustrated when they realized they had forgotten their tickets in the classroom. She tried to find a solution but ultimately had to leave the theater without seeing the play. & 3 & 3 & 3 & 3 & 5 \\
RRR\textsubscript{N} & They raced back, found tickets, and rushed in. The play was unforgettable—Amy cheered, then met the cast proudly. & 3 & 3 & 3 & 3 & 5 \\

\midrule
\multicolumn{7}{@{}l}{\textbf{Olmo}}\\
Base & Due to forgetting their tickets, Amy and her classmates couldn't enter the theater. Everyone was disappointed but decided to talk to the teachers about the situation to see if they could get in later. The experience left them eager to try again. & 3 & 3 & 1 & 1 & 5 \\
Few-Shot & Amy was frustrated that they couldn't go into the show. She wished they had remembered the tickets so they could have enjoyed the performance and maybe even spoken with the cast afterward. & 3 & 3 & 1 & 1 & 4 \\
SFT & Amy was in awe of the actors discipline and talent. After the play she was able to speak with the cast. Amy was thrilled. & 2 & 1 & 3 & 1 & 3 \\
RRR\textsubscript{O} & Amy was frustrated by not being able to enter the theater and had to explain the situation to staff before being redirected to a nearby event. Despite disappointment, she appreciated seeing live theatre from a different angle. & 3 & 3 & 3 & 3 & 5 \\
RRR\textsubscript{N} & Realising the mix-up, the group rushed back to class, scrambled for tickets, and finally boarded the bus—reassured and racing to the theater. The play began, applause echoed, and relief washed over them. & 3 & 3 & 3 & 3 & 5 \\

\midrule
\multicolumn{7}{@{}l}{\textbf{Codex}}\\
Code-Prompt & Amy was in awe of the actors discipline and talent. After the play she was able to speak with the cast. Amy was thrilled. & 1 & 1 & 3 & 1 & 3 \\

\bottomrule
\end{tabularx}

\caption{Example outputs from the test set. Ratings are given by LLM-as-judge (Gemini-3-Flash): L=Logical, R=Rational, C=Complete\textsubscript{N}, M=min\textsubscript{LRC}, N=Narrativity.}
\label{tab:consolidated-merged-amy}
\end{table}

\end{document}